\documentclass[sigconf]{acmart}
\AtBeginDocument{%
  }

\setcopyright{acmlicensed}
\copyrightyear{2018}
\acmYear{2018}
\acmDOI{XXXXXXX.XXXXXXX}

\acmConference[Conference acronym 'XX]{Make sure to enter the correct
  conference title from your rights confirmation emai}{June 03--05,
  2018}{Woodstock, NY}
\acmISBN{978-1-4503-XXXX-X/18/06}

\usepackage{array}
\usepackage{booktabs}
\usepackage{multirow}
\usepackage{subcaption}
\usepackage{graphicx}
\usepackage{algorithm}
\usepackage{algpseudocode}
\usepackage{amsthm}

\DeclareMathOperator*{\argmin}{arg\,min}

\newtheorem{proposition}{Proposition}

\begin{document}

\title{TCGU: Data-centric Graph Unlearning based on \\ Transferable Condensation}


\author{Fan Li}
\affiliation{%
 \institution{University of New South Wales}
 \city{Sydney}
 \country{Australia}}
\email{fan.li8@unsw.edu.au}

\author{Xiaoyang Wang}
\affiliation{%
  \institution{University of New South Wales}
 \city{Sydney}
 \country{Australia}}
\email{xiaoyang.wang1@unsw.edu.au}

\author{Dawei Cheng}
\affiliation{%
  \institution{Tongji University}
  \city{Shanghai}
  \country{China}}
\email{dcheng@tongji.edu.cn}

\author{Wenjie Zhang}
\affiliation{%
  \institution{University of New South Wales}
 \city{Sydney}
 \country{Australia}}
\email{wenjie.zhang@unsw.edu.au}

\author{Ying Zhang}
\affiliation{%
  \institution{University of Technology Sydney}
 \city{Sydney}
 \country{Australia}}
\email{ying.zhang@uts.edu.au}

\author{Xuemin Lin}
\affiliation{%
  \institution{Shanghai Jiao Tong University}
  \city{Shanghai}
  \country{China}}
\email{xuemin.lin@gmail.com}

\renewcommand{\shortauthors}{Trovato et al.}

\begin{abstract}
With growing demands for data privacy and model robustness, graph unlearning (GU), which erases the influence of specific data on trained GNN models, has gained significant attention.
However, existing exact unlearning methods suffer from either low efficiency or poor model performance. 
While being more utility-preserving and efficient, current approximate unlearning methods are not applicable in the zero-glance privacy setting, where the deleted samples cannot be accessed during unlearning due to immediate deletion requested by regulations. Besides, these approximate methods, which try to directly perturb model parameters still involve high privacy concerns in practice. To fill the gap, 
we propose Transferable Condensation Graph Unlearning (TCGU), a data-centric solution to zero-glance graph unlearning. Specifically, we first design a two-level alignment strategy to pre-condense the original graph into a small yet utility-preserving dataset.
Upon receiving an unlearning request, we fine-tune the pre-condensed data with a low-rank plugin, to directly align its distribution with the remaining graph, thus efficiently revoking the information of deleted data without accessing them. A novel similarity distribution matching approach and a discrimination regularizer are proposed to effectively transfer condensed data and preserve its utility in GNN training, respectively. Finally, we retrain the GNN on the transferred condensed data. 
Extensive experiments on 6 benchmark datasets demonstrate that TCGU can achieve superior performance in terms of model utility, unlearning efficiency, and unlearning efficacy than existing GU methods.
\end{abstract}

\maketitle

\section{Introduction}
\label{intro}

Graph is a ubiquitous data structure that can model the complex interactions among entities in various domains, such as finance~\cite{cheng2023anti}, biotechnology~\cite{ma2023single}, and social analysis~\cite{zhou2023hierarchical}. To exploit the rich information in graphs, a new family of machine learning models, graph neural networks (GNNs), has been proposed and has achieved great success in many graph-related tasks ~\cite{kipf2016semi,luo2024cross,wu2019net}. However, due to the increasing concern about privacy-sensitive or unreliable data elements, we are often requested to remove a specific set of graph data (e.g., nodes, edges, and their features) and its impact on the pre-trained GNN model in practice. For example, “the right to be forgotten”, which allows individuals to request the deletion of their data, has been included in many regulations, such as the European Union’s General Data Protection Regulation (GDPR)~\cite{gdpr} and the California Consumer Privacy Act (CCPA)~\cite{ccpa}. Besides, removing the influence of industry-related noise~\cite{li2024towards} and poisoned data~\cite{zhang2020gnnguard} from a model will also enhance model robustness. Thus, it is crucial to effectively handle information deletion from trained GNNs, a process known as graph unlearning (GU).

Existing GU methods can be roughly classified into two categories: exact unlearning~\cite{chen2022graph,wang2023inductive} and approximate unlearning~\cite{cheng2022gnndelete,wu2023gif,li2024towards}. The former aims to exactly obtain the model parameters trained without the deleted data. A naive way is to retrain the model from scratch on the remaining graph. However, this approach is resource-consuming, especially on large-scale networks. To deal with this, GraphEraser~\cite{chen2022graph} proposes a shard-based framework that only retrains the sub-model on shards with deleted data and aggregates the results. Following this architecture, GUIDE~\cite{wang2023inductive} further optimizes the graph partition and aggregation methods to enhance model performance. While shard-based methods ensure the removal of deletion targets, the partition severely destroys the graph topology, which is vital for GNN training, hence hurting the predictive performance of the unlearned model.


To achieve a better tradeoff between model utility and unlearning efficiency, a line of approximate unlearning methods has been developed. For example, GIF~\cite{wu2023gif} extends influence function (IF)~\cite{koh2017understanding} to graph learning and approximately estimates weight modification caused by removed data. MEGU~\cite{li2024towards} proposes a mutual evolution framework for predictive and unlearning modules, achieving superior model performance. However, there are two key limitations in recent approximate methods: (1) They cannot handle the zero-glance setting~\cite{tarun2023fast,cong2023efficiently}, where deletion targets are removed immediately due to strict regulatory compliance, and cannot be accessible even for the unlearning purpose. This setting offers a higher level of privacy guarantees. (2) Empirical results (Section~\ref{unlearn_efficacy}) show that they fail to achieve satisfactory performance in removing the influence of deleted targets. For instance, they are susceptible to membership inference attacks, leading to higher privacy concerns. 


We notice that the current approximate methods are primarily model-centric, which perturb the parameters of the trained model with well-designed objectives based on prior deleted samples. For instance, GIF estimates parameter change with graph-oriented IF utilizing deletion targets. In contrast, exact methods directly eliminate the information of deletion requests from a data perspective, which is more effective and follows the zero-glance setting. This motivates us to investigate GU from the view of data to achieve more efficient and utility-preserving retraining.
In recent years, graph condensation (GC)~\cite{jin2021graph,zheng2023structure,wang2024fast,fang2024exgc}, which aims to distill the original large graph into a small yet informative network, has become a highly effective data-centric solution for efficient GNN training. If we can simply erase the information of deleted targets compressed in the small-scale data similar to the exact methods, the retraining efficiency could be facilitated and model utility could be guaranteed. Different from current approximate methods, this can leave us from designing implicit parameter update objectives and make us focus on more direct and fundamental data-level information removal, which may improves forgetting power.






Inspired by this, we introduce TCGU, a data-centric graph unlearning method based on transferable data condensation. Specifically, our pipeline consists of three stages: pre-condensation, condensed data transfer, and model retraining. First, we adopt a two-level alignment graph condensation (TAGC) method, which comprises covariance-preserving feature-level alignment (CFA) and logits-level alignment (LA), to distill a small-scale synthetic graph to replace the original large one, ensuring GNN models trained on both datasets have comparable performance. After the arrival of an unlearning request, instead of recondensing the remaining graph which is time-consuming, we propose to fine-tune the pre-condensed graph with a low-rank residual plugin to align its distribution with the remaining graph. A novel similarity distribution matching (SDM) technique is also proposed to match inter-class relationships between the remaining and condensed graphs. In this way, we unlearn the knowledge of deletion targets from condensed data efficiently without reliance on them.
To further preserve data utility in the transfer stage, we apply a contrastive-based regularizer to distinguish features extracted from condensed data. Finally, we retrain the GNN on the updated condensed graph and obtain the unlearned model. We conduct extensive experiments on 6 benchmark datasets with 4 popular GNN backbones. The results show that TCGU achieves a better trade-off in model utility and unlearning efficiency without prior information on deleted data. Furthermore, its superior performance in efficacy evaluation with advanced membership inference attack (MIA)~\cite{chen2021machine} and edge attack (EA)~\cite{wu2023gif} validates its effectiveness in data removal. 
The main contributions of the paper are summarized as follows.
\begin{itemize}
    \item We propose TCGU, a novel data-centric graph unlearning framework, to achieve efficient, effective, and utility-preserving unlearning in the zero-glance setting. To the best knowledge, we are the first data-centric approximate GU method that can deal with immediate data deletion.
    \item To distill the large original graph into smaller yet informative data, we introduce a two-level alignment strategy during pre-processing. Additionally, a fine-tuning strategy that combines a low-rank plugin with a similarity distribution matching technique is employed to efficiently align the distribution of the pre-condensed data with that of the remaining graph. Furthermore, a contrastive-based discrimination regularizer is incorporated into the data transfer process to preserve data utility.
    \item Extensive experiments on 6 benchmark datasets and 4 representative GNNs demonstrate that our TCGU achieves a better trade-off between efficiency and model utility under the zero-glance setting. The comprehensive unlearning efficacy test verifies its capability in data deletion.
\end{itemize}

\section{Related Work}
\label{rel}

With the increasing need for data privacy, machine unlearning (MU), which aims to rule out the influence of the data from the trained ML models, has become a trending topic~\cite{bourtoule2021machine,cao2015towards,chundawat2023can,guo2020certified,jia2023model}. Graph unlearning refers to machine unlearning on graph-structured data.
In this paper, we focus on GNN models. Although training from scratch is straightforward, this strategy is inefficient on large datasets. To reduce time cost in exact unlearning, GraphEraser~\cite{chen2022graph} proposes a shard-based unlearning framework that utilizes graph partition to facilitate efficient retraining. Built upon this architecture, GUIDE~\cite{wang2023inductive} designs a fairness-aware partition algorithm and a similarity-based aggregation strategy to improve model utility. However, the partition would inevitably impair graph topology, which is essential to model performance. 
Projector~\cite{cong2023efficiently} and GraphEditor~\cite{cong2022grapheditor} provide closed-form solutions with information deletion guarantees. However, these two methods are only applicable to linear GNNs and cannot handle large deletion requests. To achieve a better trade-off in efficiency and utility, many approximate graph unlearning methods have been proposed. GIF~\cite{wu2023gif} design a graph-oriented
influence function that considers the structural influence of deleted
entities on their neighbors. GNNDelete~\cite{cheng2022gnndelete} introduces a layer-wise operator to remove the influence of deleted elements while preserving the remaining model knowledge. MEGU~\cite{li2024towards} achieves mutual benefits between predictive and unlearning modules in a unified optimization framework. However, these approximate strategies are still prone to extra information leakage and they require information on the deleted targets to achieve unlearning, which may be impractical in real-world scenarios~\cite{suriyakumar2022algorithms,tarun2023fast}.

\section{Preliminary}
\label{pre}


Consider an attributed graph \(\mathcal{G}(\mathcal{V},\mathcal{E},\mathcal{X})\) with \(|\mathcal{V}|=N\) nodes, \(|\mathcal{E}|=M\) edges, and \(\mathcal{X}=\mathbf{X}\). Each node \(v_{i}\) is associated with a \(F\)-dimensional feature vector \(x_{i} \in \mathcal{X}\), where \(\mathcal{X} \in \mathbb{R}^{N \times F}\) is feature matrix. \(\mathbf{A},\mathbf{D} \in \mathbb{R}^{N \times N}\) denote the adjacency matrix and its associated degree matrix. We use \(\mathbf{Y} \in \mathbb{R}^{N \times C}\) to denote the \(C\)-classes of node labels, where $y_{i} \in \mathbf{Y}$ is the one-hot label vector for node \(v_{i}\). 


\subsection{Background Knowledge}


\noindent\textbf{Graph condensation (GC)} As an extension of data condensation~\cite{wang2022cafe,zhao2021dataset,zhao2023dataset} in the graph domain, graph condensation~\cite{jin2021graph,jin2022condensing,zheng2023structure,xiao2024} aims to learn a downsized synthetic graph \(\mathcal{G}^{\prime}(\mathcal{V}^{\prime},\mathcal{E}^{\prime},\mathcal{X}^{\prime})\) with \(|\mathcal{V}^{\prime}|=N^{\prime} \ll N\) from original graph \(\mathcal{G}\), such that a GNN trained on \(\mathcal{G}^{\prime}\) can achieve comparable performance to one trained on \(\mathcal{G}\). Formally, GC can be defined as solving the following problem:
\begin{equation}
\begin{split}
    \min_{\mathcal{G^{\prime}}} & \mathcal{L}(\mathrm{GNN_{\theta_{\mathcal{G}^{\prime}}}}(\mathbf{A},\mathbf{X}),\mathbf{Y}) \\
    \text{s.t.}  \quad  \theta_{\mathcal{G}^{\prime}} = &\argmin_{\theta} \mathcal{L}(\mathrm{GNN}_{\theta}(\mathbf{A}^{\prime},\mathbf{X}^{\prime}),\mathbf{Y}^{\prime}),
\end{split}
\end{equation}
where \(\mathbf{A}^{\prime} \in \mathbb{R}^{N^{\prime} \times N^{\prime}}, \mathbf{X}^{\prime} \in \mathbb{R}^{N^{\prime} \times F}, \mathbf{Y}^{\prime} \in \mathbb{R}^{N^{\prime} \times C}\) denote adjacency matrix, feature matrix, and label matrix for the condensed graph, respectively. GC has emerged as a popular data-centric solution for efficient GNN training and also shows promise for various other applications, such as neural architecture search (NAS)~\cite{jin2021graph}, privacy preservation~\cite{pan2023fedgkd}, and continual learning~\cite{liu2023cat}.

\vspace{1mm} 
\noindent \textbf{Distribution matching (DM).}  Distribution Matching is a recently proposed data condensation method~\cite{yuan2023real,zhao2023dataset,zhao2023improved,zheng2023rdm}. Given a target dataset \(\mathcal{R}=\{(x_{i},y_{i})\}_{i=1}^{|\mathcal{R}|}\) and a synthetic dataset \(\mathcal{S}=\{(s_{j},y_{j}^{s})\}_{j=1}^{|\mathcal{S}|}\), DM aims to minimize the maximum mean discrepancy (MMD)~\cite{gretton2012kernel} between target data distribution \(q(x)\) and synthetic data distribution \(p_{_{\phi}}(s)\) in embedding space as:
\begin{equation}
\label{e1}
    \underset{\Vert \psi_{\theta} \Vert_{\mathcal{H}} \leq 1}{\sup} (\mathbb{E}_{q}[\psi_{\theta}(\mathcal{R})]-\mathbb{E}_{p}[\psi_{\theta}(\mathcal{S})]),
\end{equation}
where \(\psi_{\theta}\) is an embedding function residing within a unit ball in the universal Reproducing Kernel Hilbert Space \(\mathcal{H}\) (RKHS)~\cite{hilbert1906grundzuge}. Existing methods adopt neural network as \(\psi_{\theta}\) and simplify Eq.~\ref{e1} by empirically
estimating the expectations for all distributions as:
\begin{equation}
    \min_{\mathcal{S}} \mathbb{E}_{\theta \sim \mathcal{P}(\theta)} \Vert \frac{1}{|\mathcal{R}|}\sum_{i=1}^{|\mathcal{R}|}(\psi_{\theta}(x_{i}))-\frac{1}{|\mathcal{S}|}\sum_{j=1}^{|\mathcal{S}|}(\psi_{\theta}(s_{j}))\Vert
\end{equation}
where \(\mathcal{P}(\theta)\) denotes the distribution of model parameters. Most prior works~\cite{liu2022graph,zhao2023dataset,zheng2023rdm} rely on random initialization of \(\theta \sim \mathcal{P}(\theta)\), which disregards pre-training and improves efficiency.

\subsection{Problem Definition}

Following~\cite{tarun2023fast}, we formulate the graph unlearning problem in zero-glance privacy settings. Given the original GNN model \(\mathcal{F}_{o}\) trained on graph \(\mathcal{G}\) and the unlearning request \(\Delta \mathcal{G}=\{\Delta \mathcal{V},\Delta \mathcal{E},\Delta \mathcal{X}\}\). The goal of zero-glance graph unlearning is to find a mechanism \(\mathcal{M}\) that only takes \(\mathcal{F}_{o}\) and the remaining graph \(\mathcal{G} \backslash \Delta \mathcal{G}\) as inputs, and then outputs an unlearned model \(\mathcal{F}_{u}\) that is approximately the same as the GNN model \(\mathcal{F}_{\mathcal{G} \backslash \Delta \mathcal{G}}\) obtained by retraining from scratch on \(\mathcal{G} \backslash \Delta \mathcal{G}\). The objective can be formulated as:
\begin{equation}
    \begin{split}
             \min & \mathcal{D}(\mathcal{F}_{u},\mathcal{F}_{\mathcal{G} \backslash \Delta \mathcal{G}}) \\
             \mathrm{s.t.} \mathcal{F}_{u} & =\mathcal{M}(\mathcal{F}_{o},\mathcal{G} \backslash \Delta \mathcal{G}),
    \end{split}
\end{equation}
where \(\mathcal{D}(\cdot,\cdot)\) is the discrepancy function. Different from 
 the setting in prior approximate methods~\cite{wu2023gif,cheng2022gnndelete,li2024towards}, we cannot access the forget set data \(\Delta \mathcal{G}\) during unlearning.



\section{Methodology}
\label{method}

\subsection{Overviw of TCGU Framework}

\begin{figure*}[t]
\centering
\includegraphics[width=1\textwidth]{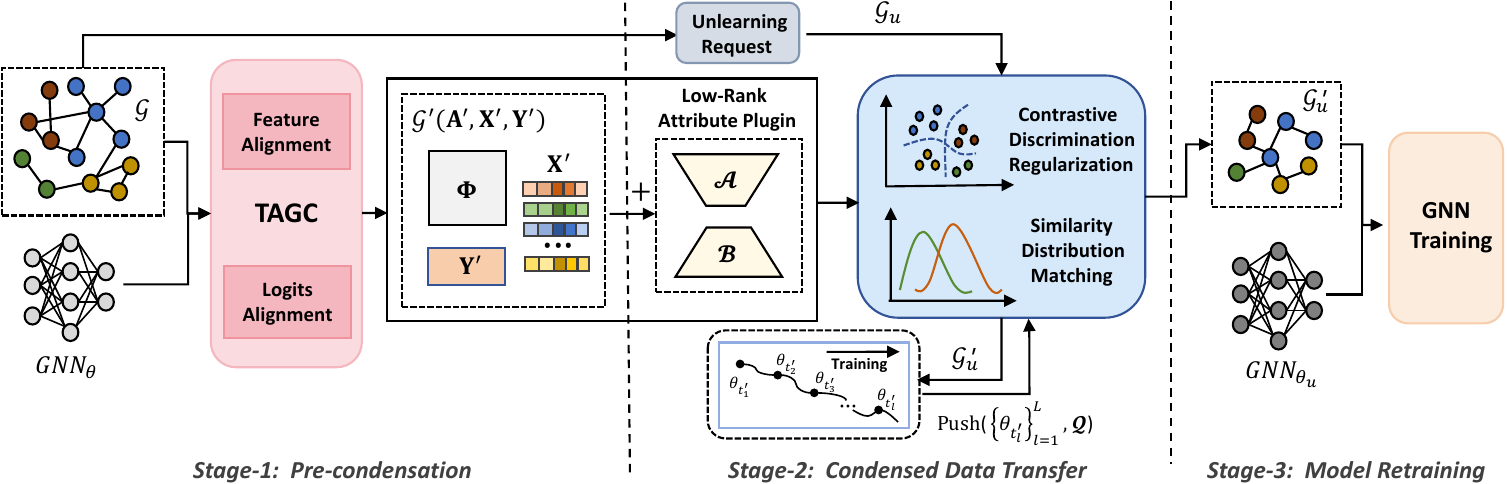} 
\caption{The overall pipeline of TCGU}
\label{fig:arch}
\end{figure*}

As shown in Fig.~\ref{fig:arch}, TCGU consists of three stages: pre-condensation, condensed data transfer, and model retraining. In stage 1, we design a two-level alignment graph condensation (TAGC) method to distill the original graph \(\mathcal{G}\) into a small yet informative \(\mathcal{G}^{\prime}\) which preserves training knowledge of \(\mathcal{G}\). After the arrival of unlearning requests, instead of directly retraining the model on the remaining large graph \(\mathcal{G}_{r}\) which is computationally expensive, we propose to fine-tune the pre-condensed graph with a residual low-rank plugin so that the condensed data can be efficiently transferred into the distribution of \(\mathcal{G}_{r}\), removing the influence of deleted targets from data-centric perspective. More specifically, 
we devise a similarity distribution matching (SDM) technique together with the feature-level alignment, to match the distribution of the condensed graph and the remaining graph from inter and intra-class relationships without recondensation. In SDM, we sample the embedding function with a novel trajectory-based method, allowing for significantly diversifying the parameter distribution in MMD computation. To preserve the utility of transferred condensed data in GNN training, we propose a contrastive-based discrimination regularizer to make the extracted features in the synthetic graph more distinguishable. Note that we only rely on pre-condensed data and \(\mathcal{G}_{r}\) in the transfer stage, satisfying the zero-glance setting. Finally, we retrain the GNN on the small transferred data and obtain the unlearned model.

\subsection{Two-level Alignment Graph Condensation}
\label{cond}

In this subsection, we present our pre-condensation algorithm based on two-level alignment. Firstly, we give the modeling strategy of the learnable condensed data. Then, we propose covariance-preserving feature alignment (CFA) to align both structure and node attributes between \(\mathcal{G}\) and \(\mathcal{G^{\prime}}\), and discuss its relationship with the popular gradient matching objective in graph condensation. After that, to further distill GNN training knowledge into synthetic data, we develop logits alignment (LA) based on GNN to be unlearned. 

\vspace{1mm} 
\noindent \textbf{Condensed Data Modeling.} To model the learnable condensed graph \(\mathcal{G}^{\prime}(\mathbf{A}^{\prime},\mathbf{X}^{\prime},\mathbf{Y}^{\prime})\), 
we treat feature matrix \(\mathbf{X}^{\prime}\) as free parameter and parameterize \(\mathbf{A}^{\prime}\) as a function of the condensed features as:
\begin{equation}
    \mathbf{A}^{\prime}=g_{\Phi}(\mathbf{X}^{\prime}), \text{with } \mathbf{A}^{\prime}_{ij} = \sigma\Big(\frac{\mathbf{MLP}_{\Phi}([x_{i};x_{j}])+\mathbf{MLP}_{\Phi}([x_{j};x_{i}])}{2}\Big)
\end{equation}
where \(\mathbf{MLP}_{\Phi}\) is a multi-layer perceptron parameterized with \(\Phi\) and \([\cdot;\cdot]\) denotes concatenation. \(\sigma(\cdot)\) represents \(\mathrm{Sigmoid}(\cdot)\) activation function. This design can effectively capture the implicit relationship between graph topology and node features. For node labels \(\mathbf{Y}^{\prime}\), we keep the class distribution the same as the original training graph \(\mathbf{Y}\) followed by~\cite{jin2021graph}. With this parameterization, we can only optimize \(\mathbf{X}^{\prime}\) and \(\Phi\) during condensation.

\vspace{1mm} 
\noindent \textbf{Covariance-Preserving Feature Alignment.}
To jointly encode the graph structure and node attributes, we first perform feature propagation on \(\mathcal{G}\) and obtain multi-hop feature representations as:
\begin{equation}
\begin{split}
    & \mathbf{H}^{(k)} =\mathbf{P}^{k}\mathbf{X}, k=0,1,...K \\
    \mathbf{P} = (w_{loop}&\mathbf{I}+\mathbf{D})^{-\frac{1}{2}}(w_{loop}\mathbf{I}+\mathbf{A})(w_{loop}\mathbf{I}+\mathbf{D})^{-\frac{1}{2}},
\end{split}
\end{equation}
where \(\mathbf{H}^{(k)}\) is the \(k\)-th hop feature representations and \(\mathbf{H}^{(0)}=\mathbf{X}\). \(K\) is the hop number. \(\mathbf{P}\) denotes the propagation matrix and \(w_{loop} \in \mathbb{R}^{+}\) is the self-loop weight. This is a general transition matrix that is equivalent to performing a lazy random walk with a probability of staying at node \(v_{i}\) of \(p_{stay, i}=\frac{w_{loop}}{\mathbf{D}_{i}}\). Similarly, we could obtain feature representations \(\mathbf{H}^{\prime (k)}\) for condensed graph \(\mathcal{G}^{\prime}\). To align two graphs at the feature level, we propose to satisfy two properties: mean feature alignment and feature covariance alignment. For the first property, within each class \(c\), we minimize the distance between the average node representations of \(\mathcal{G}\) and \(\mathcal{G}^{\prime}\) in a hierarchical way as:
\begin{equation}
    \mathcal{L}_{mean} = \sum_{k=0}^{K}\sum_{c=1}^{C}r_{c} \cdot \big\Vert \mu_{c}^{(k)}-\mu_{c}^{\prime (k)} \big\Vert_{2}^{2}
\end{equation}
where \(\mathcal{L}_{mean}\) denotes the mean matching loss. \(\mu_{c}^{(k)},\mu_{c}^{\prime (k)}\) are averaged \(k\)-hop node feature vectors of class \(c\) for \(\mathcal{G}\) and \(\mathcal{G}^{\prime}\), respectively. \(r_{c}\) is node ratio of \(c\) in training graph. 
\(\big\Vert \cdot \big\Vert_{2}^{2}\) denotes L2-norm. However, simply matching the mean feature would overlook the dependency between different feature dimensions. Thus, we propose to minimize the distance of class-wise covariance matrices between \(\mathcal{G}\) and \(\mathcal{G}^{\prime}\) in a hierarchical way as:
\begin{equation}
    \begin{split}
            & \mathcal{L}_{cov}  = \sum_{k=0}^{K}\sum_{c=1}^{C}r_{c} \cdot \big\Vert \mathbf{U}_{c}^{(k)}-\mathbf{U}_{c}^{\prime (k)} \big\Vert_{2}^{2} \\
            \mathbf{U}_{c}^{(k)} & = \frac{1}{N_{c}-1}\Tilde{\mathbf{H}}_{c}^{(k) T}\Tilde{\mathbf{H}}_{c}^{(k)}, \quad \Tilde{\mathbf{H}}_{c}^{(k)}={\mathbf{H}}_{c}^{(k)}-\mu_{c}^{(k)} \\
            \mathbf{U}_{c}^{\prime (k)} & = \frac{1}{N_{c}^{\prime}-1}\Tilde{\mathbf{H}}_{c}^{\prime (k) T}\Tilde{\mathbf{H}}_{c}^{\prime (k)}, \quad\Tilde{\mathbf{H}}_{c}^{\prime (k)}={\mathbf{H}}_{c}^{\prime (k)}-\mu_{c}^{\prime (k)}
    \end{split}
\end{equation}
where \(\mathcal{L}_{cov}\) denotes covariance matching loss. \(\mathbf{U}_{c}^{(k)}\) and \(\mathbf{U}_{c}^{\prime (k)}\) denote the \(k\)-th hop feature covariance matrices of class \(c\) for \(\mathcal{G}\) and \(\mathcal{G}^{\prime}\), respectively. \(\mathbf{H}_{c}^{(k)} \in \mathbb{R}^{N_{c} \times F}\) and \(\mathbf{H}_{c}^{\prime (k)} \in \mathbb{R}^{N_{c}^{\prime} \times F}\) are the corresponding feature matrices of class \(c\) . \(N_{c}\) and \(N_{c}^{\prime}\) represent the number of nodes in class \(c\) for \(\mathcal{G}\) and \(\mathcal{G}^{\prime}\), respectively. We aggregate the above two objectives and obtain feature alignment loss \(\mathcal{L}_{feat}\): 
\begin{equation}
\label{feat_loss}
    \mathcal{L}_{feat} = \mathcal{L}_{mean}+\lambda_{c}\mathcal{L}_{cov}
\end{equation}
where \(\lambda_{c}\) is a hyperparameter to control the weight of covariance loss. To justify the rationale of the feature-level alignment, we establish a connection between FA and gradient matching~\cite{jin2021graph}. The proof of the following proposition is deferred to Appendix C.1.
\begin{proposition}
\label{pro1}
  Covariance-preserving feature alignment objective provides an
upper bound for the gradient matching objective. 
\end{proposition}


\vspace{1mm} 
\noindent \textbf{Pre-trained GNN for Logits Alignment.} Feature-level alignment matches joint distribution of topology and attributes of \(\mathcal{G}\) and \(\mathcal{G}^{\prime}\). However, another desired property of \(\mathcal{G}^{\prime}\) is that it can preserve the GNN training performance on \(\mathcal{G}\). Thus, considering \(\mathbf{Y}\) shares a similar distribution of \(\mathbf{Y}^{\prime}\), it is also crucial to align the model performance on two graphs, which is the distribution of the output layer (i.e., logits in the classification task). As the original model \(\text{GNN}_{\theta_{\mathcal{G}}}\) trained on \(\mathcal{G}\) has minimized the training loss, it should also achieve minimized node classification loss on \(\mathcal{G^{\prime}}\) as:
\begin{equation}
    \begin{split}
         \underset{\mathbf{X}^{\prime},\Phi}{\min} \mathcal{L}_{logits} & = \mathcal{L}_{cls}(\text{GNN}_{\theta_{\mathcal{G}}}(g_{\Phi}(\mathbf{X}^{\prime}),\mathbf{X}^{\prime}),\mathbf{Y}^{\prime}) \\
         \mathrm{s.t.} \theta_{\mathcal{G}} & = \underset{\theta}{\arg\min} \mathcal{L}_{cls}(\text{GNN}_{\theta}(\mathbf{A},\mathbf{X}),\mathbf{Y}),
    \end{split}
\end{equation}
where \(\mathcal{L}_{logits}\) denotes logits alignment loss. \(\mathcal{L}_{cls}\) represents the classification loss function used to measure the difference between model predictions and ground truth (i.e. cross-entropy loss). We point out that this logits alignment is similar to knowledge distillation, and the knowledge is transferred from original data \(\mathcal{G}\) (i.e., teacher) to the synthetic data \(\mathcal{G}^{\prime}\) (i.e., student). Moreover, this process preserves inductive bias of the unlearned GNN architecture in the condensed data, making models trained on both the condensed and original data remain closer in parameter space.

\vspace{1mm} 
\noindent \textbf{Condensation Optimization.} 
During condensation, we jointly optimize the feature alignment loss and logits alignment loss as:
\begin{equation}
    \underset{\mathbf{X}^{\prime},\Phi}{\min} \mathcal{L}_{cond} = \mathcal{L}_{logits} + \lambda_{f}\mathcal{L}_{feat},
\end{equation}
where \(\mathcal{L}_{cond}\) is the overall condensation loss and \(\lambda_{f}\) denotes coefficient of feature alignment loss. As \(\mathbf{X}^{\prime}\) and \(\Phi\) are interdependent, jointly optimizing them is difficult. To tackle this, we alternatively update \(\mathbf{X}^{\prime}\) and \(\Phi\) during condensation.
Finally, we use \(\mathbf{A}^{\prime}=\mathrm{RELU}(g_{\Phi}(\mathbf{X}^{\prime})-\delta)\) to obtain the sparsified graph structure, where \(\delta\) is the threshold. The overall condensation algorithm is shown in Algorithm~\ref{alg:gc} in Appendix~\ref{method_supp}.

\vspace{1mm} 
\subsection{Efficient Fine-tuning for Data Transfer}
\label{finetune}

Upon receiving a deletion request,
directly condensing the \(\mathcal{G}_{r}\) with the proposed TAGC is intractable as (1) condensing the graph from scratch upon each unlearning request is inefficient. (2) logits alignment with \(\text{GNN}_{\theta_{\mathcal{G}}}\) may cause synthesis bias as the distribution discrepancy between \(\mathcal{G}\) and \(\mathcal{G}_{r}\) may lead to sub-optimal performance for \(\text{GNN}_{\theta_{\mathcal{G}}}\) on \(\mathcal{G}_{r}\) (e.g., \(\mathcal{G}\) is under adversarial attack).
To address this, with pre-condensed \(\mathcal{G}^{\prime}\) that contains distribution and training knowledge of \(\mathcal{G}\), we propose to fine-tune this small-scale data to align its distribution with that of \(\mathcal{G}_{r}\), which equals to remove the influence of deleted data on \(\mathcal{G}^{\prime}\). The proposed data transfer method consists of three key techniques including a low-rank knowledge transfer plugin, similarity distribution matching, and a contrastive-based discrimination regularizer.


\vspace{1mm} 
\noindent \textbf{Low-rank Plugin for Distribution Transfer.}
Inspired by the low-rank structure~\cite{hu2021lora,pope2020intrinsic}, which is prevalent in domain adaptations, we propose to transfer the graph distribution knowledge in a low-rank manifold. Specifically, we first freeze the pre-trained condensed node features \(\mathbf{X}^{\prime} \in \mathbb{R}^{N^{\prime} \times F}\) so that it would not receive gradient updates in the fine-tuning stage. Then, we initialize two learnable low-rank matrices \(\mathcal{A} \in \mathbb{R}^{N^{\prime} \times r}\) and \(\mathcal{B} \in \mathbb{R}^{r \times F}\), where \(r \ll \min\{N^{\prime},F\}\). During fine-tuning, the updated node feature \(\mathbf{X}^{\prime}_{u}\) and graph structure \(\mathbf{A}^{\prime}_{u}\) can be formulated as:
\begin{equation}
    \begin{split}
        \mathbf{X}^{\prime}_{u} = \mathbf{X}^{\prime} & + \Delta \mathbf{X}^{\prime} = \mathbf{X}^{\prime} + \mathcal{A}\mathcal{B} \\
        & \mathbf{A}^{\prime}_{u} = g_{\Phi}(\mathbf{X}^{\prime}_{u}),
    \end{split}
\end{equation}
where \(\Delta \mathbf{X}^{\prime}\) is the transfer block of features. We use a random gaussian initialization for \(\mathcal{A}\) and zero for \(\mathcal{B}\), so \(\Delta \mathbf{X}^{\prime}=\mathcal{A}\mathcal{B}\) is zero at the beginning of fine-tuning. We analyze that this low-rank plugin (LoRP) has two desired properties in our distribution transfer process. (1) The computation cost in fine-tuning is significantly reduced as trainable parameters for node features are transformed into two small low-rank matrices. (2) This decomposition-based transfer technique with a residual connection can well balance between preserving the remaining source-domain graph knowledge and flexibly accommodating the new distribution. These properties make our data transfer efficient and stable.


\vspace{1mm} 
\noindent \textbf{Similarity Distribution Matching.} The proposed feature-level alignment only considers matching the intra-class data distribution while ignoring matching the inter-class relationship between two graphs, which is important in distribution alignment. Logits alignment complements this by capturing the difference between classes in $\mathcal{G}$ with the trained GNN and transferring this knowledge to the condensed graph. As LA is not applicable in the fine-tuning stage, we propose a novel strategy to transfer inter-class information of \(\mathcal{G}_{r}\) to the condensed data based on distribution matching. Different from existing DM-based condensation methods~\cite{zhao2023dataset,zheng2023rdm}, we have two key modifications: (1) trajectory-based function sampling (TFS). (2) similarity matching objective. 

\vspace{1mm} 
\noindent \underline{\textit{Trajectory-based Function Sampling.}} As discussed in Section~\ref{pre}, randomly initialized neural network is commonly used as sampled embedding function \(f_{\theta}\) in MMD computation.
However, recent work~\cite{zhao2023improved} finds this simple predefined distribution has a limited number of patterns and occupies only a small fraction of the hypothesis space. To solve this, we propose to sample \(f_{\theta}\) from the training trajectory \(\mathcal{T}\) of the model trained on the updated condensed graph \(\mathcal{G}^{\prime}_{u}\). Specifically, we initialize a parameter queue \(\mathcal{Q}\). After we fine-tune the \(\mathcal{G}^{\prime}_{u}\) for every \(\tau_{s}\) steps, we train a random initialized GNN on \(\mathcal{G}^{\prime}_{u}\) for \(T_{s}\) epochs. In this training trajectory, we sample \(L_{s}\) model with at equal epoch intervals and push intermediate model parameters \(\{\theta_{t^{\prime}_{l}}\}_{l=1}^{L_{s}}\) into \(\mathcal{Q}\), where \(\theta_{t^{\prime}_{l}}\) denotes the \(l\)-th sampled model parameters in the trajectory. The evolving training trajectories are treated as our function distribution \(\mathcal{P}(\theta)\). We note that this technique allows for a significant diversification of the parameter distribution so that more informative features can be extracted in distribution transfer. Besides, sampling multiple intermediate parameters in the training trajectory of a small synthetic graph is highly efficient. The detailed description of the function sampling strategy is illustrated in Algorithm~\ref{alg:ft} in Appendix B.

\vspace{1mm} 
\noindent \underline{\textit{Similarity Matching Objective.}} In each fine-tuning step, we first sample a function \(f_{\theta}\) from \(\mathcal{Q}\) and encode \(\mathcal{G}_{r}\) and \(\mathcal{G}^{\prime}_{u}\) as:
\begin{equation}
    \mathbf{Z}_{r} = f_{\theta}(\mathbf{A}_{r},\mathbf{X}_{r}), \quad \mathbf{Z}^{\prime}_{u} = f_{\theta}(g_{\Phi}(\mathbf{X}^{\prime}_{u}),\mathbf{X}_{u}),
\end{equation}
where \(\mathbf{Z}_{r} \in \mathbb{R}^{N_{r} \times d_{h}}, \mathbf{Z}^{\prime}_{u} \in \mathbb{R}^{N^{\prime} \times d_{h}}\) are embedding matrices and \(d_{h}\) is hidden dimension. To represent class-wise patterns, we introduce \textbf{prototype vector} \(\mathbf{p}_{c}\) for each class \(c\) as:
\begin{equation}
    \mathbf{p}_{c} = \frac{1}{N_{c}}\sum_{i=1}^{N_{c}} \mathbf{z}_{c,i}, \quad c=1,2,...,C
\end{equation}
where \(\mathbf{z}_{c,i}\) denotes node embedding which belongs to \(c\) and \(N_{c}\) is the number of embeddings in class \(c\). After that, to capture the inter-class relationship in graph data, we propose \textbf{similarity embedding} \(\mathbf{s}_{i}\) for each node \(v_{i}\) as:
\begin{equation}
\label{sim}
    \begin{split}
        \mathbf{s}_{i} = \big[\mathrm{sim}(\mathbf{z}_{i},\mathbf{p}_{1}) ,\mathrm{sim}(\mathbf{z}_{i},\mathbf{p}_{2}),...,\mathrm{sim}&(\mathbf{z}_{i},\mathbf{p}_{C})\big]^{T}, \\
        \mathrm{sim}(\mathbf{z}_{i},\mathbf{p}_{c}) = \mathrm{exp}\Big(\frac{\mathrm{cos}(\mathbf{z}_{i},\mathbf{p}_{c})}{\tau_{sim}}\Big). \\
    \end{split}
\end{equation}
where \(z_{i}\) is node embedding for \(v_{i}\). \(\mathrm{cos}(\cdot,\cdot)\) denotes cosine similarity and \(\tau_{sim}\) is a temperature coefficient. This design makes estimation for similarity distribution highly efficient, as we can obtain inter-class similarity for each node using \(C\) representative embeddings. Based on Eq.~\ref{sim}, we calculate similarity embedding matrices for \(\mathcal{G}_{r}\) and \(\mathcal{G}^{\prime}_{u}\) as \(\mathbf{S}_{r} \in \mathbb{R}^{N_{r} \times C}\) and \(\mathbf{S^{\prime}_{u}} \in \mathbb{R}^{N^{\prime} \times C}\), respectively. To minimize the empirical MMD estimation between the similarity distributions of two datasets, we define the similarity distribution matching loss \(\mathcal{L}_{sdm}\) during fine-tuning as:
\begin{equation}
\label{sim_loss}
        \mathcal{L}_{sdm} = \sum_{c=1}^{C} \hat{r}_{c} \cdot \big\Vert \frac{1}{N_{r,c}}\sum_{i=1}^{N_{r,c}}\mathbf{s}_{r,c,i}-\frac{1}{N_{c}^{\prime}}\sum_{i=1}^{N_{c}^{\prime}}\mathbf{s}^{\prime}_{u,c,i} \big\Vert_{2}^{2} \\
\end{equation}
where \(\mathbf{s}_{r,c,i},\mathbf{s}_{u,c,i}^{\prime}\) are similarity embeddings of class \(c\) for \(\mathcal{G}_{r}\) and \(\mathcal{G}^{\prime}_{u}\), respectively. \(N_{r,c},N_{c}^{\prime}\) represent number of nodes in class \(c\) on two graphs. \(\hat{r}_{c}\) denotes the ratio of \(c\)-class nodes in \(\mathcal{G}_{r}\). 

\vspace{1mm} 
\noindent \textbf{Contrastive-based Discrimination Regularizer (CDR).}
An advantage of logits alignment is that when it transfers learned class-wise patterns from \(\mathcal{G}\) to \(\mathcal{G}^{\prime}\), 
condensed nodes in different classes could be discriminated in embedding space, which facilitate GNN training. As SDM match the inter-class relationship between \(\mathcal{G}_{r}\) and \(\mathcal{G}^{\prime}_{u}\) at the mean-embedding level, the extracted class-wise features from the condensed graph may not be easily discriminated after transferring, which could impair the utility of synthetic data. To address this, we design a discrimination regularizer based on supervised contrastive learning:
\begin{equation}
\label{cdr}
    \mathcal{L}_{cdr} = -\sum_{v_{u,i}^{\prime} \in \mathcal{V}^{\prime}_{u}}\frac{1}{|S(i)|}\sum_{v_{u,p}^{\prime} \in S(i)}\frac{\mathrm{exp}(\mathrm{cos}(\mathbf{z}_{i},\mathbf{z}_{p})/\tau_{r})}{\sum_{v_{u,q}^{\prime} \in \mathcal{V}_{u}^{\prime}}\mathrm{exp}(\mathrm{cos}(\mathbf{z}_{i},\mathbf{z}_{q})/\tau_{r})},
\end{equation}
where \(S(i)=\{v_{u,j}^{\prime}|y_{u,j}^{\prime}=y_{u, i}^{\prime}\}\) denotes the node set that has the same label with \(v_{u, i}^{\prime}\) in \(\mathcal{G}_{u}^{\prime}\), and \(|S(i)|\) is its cardinality. \(\tau_{r}\) is the temperature coefficient for the regularizer. With this regularizer, condensed nodes from diverse classes can be better discriminated by GNN model. We optimize this objective with different embedding functions so that it can capture general discriminative patterns.

\vspace{1mm} 
\noindent \textbf{Fine-tuning Training.} 
In the fine-tuning stage, we jointly optimize feature alignment loss, similarity matching loss, and discrimination regularization loss for condensed data transfer as:
\begin{equation}
    \underset{\mathcal{A},\mathcal{B},\Phi}{\min} \mathcal{L}_{ft} = \mathcal{L}_{sm} + \lambda_{f}^{\prime}\mathcal{L}_{feat}+\lambda_{r}^{\prime}\mathcal{L}_{cdr},
\end{equation}
where \(\mathcal{L}_{ft}\) is the overall fine-tuning loss. \(\lambda_{f}^{\prime},\lambda_{r}^{\prime}\) are coefficients for \(\mathcal{L}_{feat}\) and \(\mathcal{L}_{cdr}\), respectively. For simplicity, we fix the label of unlearned data and optimize \(\mathcal{A}\), \(\mathcal{B}\), and \(\Phi\) in fine-tuning. The alternative optimization mentioned in Section~\ref{cond} is also applied. We provide detailed fine-tuning process in Algorithm~\ref{alg:ft} in Appendix~\ref{method_supp}. 

\subsection{Model Retraining on Transferred Data}

After obtaining \(\mathcal{G}^{\prime}_{u}\), which is the transferred small-scale data that aligns the distribution of \(\mathcal{G}_{r}\), we retrain the GNN model with the same architecture as \(\text{GNN}_{\theta_{\mathcal{G}}}\) to obtain the unlearned model as:
\begin{equation}
    \theta_{u}  = \underset{\theta}{\arg\min} \mathcal{L}_{cls}(\text{GNN}_{\theta}(\mathbf{A}^{\prime}_{u},\mathbf{X}^{\prime}_{u}),\mathbf{Y}^{\prime}_{u})
\end{equation}
where \(\theta_{u}\) denotes parameters of the unlearned model. The process is highly efficient as the condensed data could significantly reduce computational costs in model training.

\section{Experiments}
\label{exp}


\subsection{Experimental Setup}

\begin{table*}[th]
\caption{Overall performance on model utility and unlearning efficiency for node unlearning. We report F1 scores (F1) and running time (RT) for different methods 
}
\label{utility}
\centering
\resizebox{\linewidth}{!}{\begin{tabular}{cccccccccccccc}
\toprule
\multirow{2}{*}{GNN} & \multicolumn{1}{c|}{\multirow{2}{*}{Component}} & \multicolumn{2}{c}{Cora} & \multicolumn{2}{c}{Citeseer} & \multicolumn{2}{c}{Pubmed} & \multicolumn{2}{c}{Photo} & \multicolumn{2}{c}{Computers} & \multicolumn{2}{c}{Arxiv} \\
                     & \multicolumn{1}{c|}{}                        & F1        & RT(s)  & F1        & RT(s) & F1        & RT(s)   & F1        & RT(s) & F1        & RT(s) & F1        & RT(s)  \\ \midrule
\multirow{9}{*}{GCN}    &    \multicolumn{0}{c|}{Retrain}              & 81.95\(\pm\)0.15 & 9.78 & 73.29\(\pm\)0.31& 16.59 &86.46\(\pm\)0.55 & 62.26 & 90.83\(\pm\)0.37 & 43.07 & 84.42\(\pm\)0.24 & 90.91 & 68.72\(\pm\)0.19 &  395.97 \\
                        &    \multicolumn{0}{c|}{GE-BLPA}              & 67.78\(\pm\)0.01 & 7.69 & 64.72\(\pm\)0.00 & 9.10 & 71.82\(\pm\)0.01 & 25.34 & 76.17\(\pm\)0.04 & 15.85 & 73.20\(\pm\)0.02 & 23.99 & 58.89\(\pm\)0.04 & 50.28  \\
                        &    \multicolumn{0}{c|}{GE-BKEM}              & 51.48\(\pm\)0.00 & 7.88 & 56.54\(\pm\)0.03 & 9.34 & 60.27\(\pm\)0.02 & 24.12 &58.24\(\pm\)0.03&  14.67 & 69.75\(\pm\)0.05& 23.34 & 54.17\(\pm\)0.02 & 50.85 \\
                        &    \multicolumn{0}{c|}{GUIDE-Fast}           & 71.85\(\pm\)0.04 & 5.93& 68.10\(\pm\)0.01&6.23& 75.20\(\pm\)0.06& 17.89& 78.53\(\pm\)0.02& 9.42 & 73.11\(\pm\)0.03& 13.04 & OOM & OOM  \\
                        &    \multicolumn{0}{c|}{GUIDE-SR}             & 72.33\(\pm\)0.02 & 6.58 & 70.55\(\pm\)0.00& 7.21 & 75.89\(\pm\)0.03& 21.65 & 77.48\(\pm\)0.04 & 12.93 & 75.37\(\pm\)0.01& 14.35 & OOM & OOM  \\
                        &    \multicolumn{0}{c|}{GNNDelete}            & 75.27\(\pm\)0.11 & 1.26 & 72.18\(\pm\)0.20 & 1.45& 81.90\(\pm\)0.12 & 2.21 & 85.10\(\pm\)0.25 & 1.52 & 79.15\(\pm\)0.17& 3.02 & 65.12\(\pm\)0.06 & 10.54  \\
                        &    \multicolumn{0}{c|}{GIF}                  & \underline{79.67\(\pm\)0.36} & \textbf{0.39} & 72.97\(\pm\)0.42 & \textbf{0.46} & 80.76\(\pm\)0.45 & \textbf{0.65}& 86.83\(\pm\)0.39 & \textbf{0.41} & 80.27\(\pm\)0.25 & \textbf{0.60}  & OOM & OOM  \\
                              &    \multicolumn{0}{c|}{MEGU}           & 79.51\(\pm\)0.15 & \underline{0.52} & \underline{73.85\(\pm\)0.23} & \underline{0.54} & \textbf{84.37\(\pm\)0.07} & \underline{1.08}& \underline{88.65\(\pm\)0.01} & \underline{0.69} & \underline{83.18\(\pm\)0.02} & \underline{1.55} & \textbf{68.29\(\pm\)0.07}& \underline{7.80} \\
                              &    \multicolumn{0}{c|}{TCGU}            & \textbf{82.18\(\pm\)0.26} & 2.14 & \textbf{75.99\(\pm\)0.18} & 2.19 & \underline{83.77\(\pm\)0.22} & 4.40 & \textbf{89.92\(\pm\)0.61} & 1.95 & \textbf{84.76\(\pm\)0.22} & 2.76  & \underline{66.37\(\pm\)0.78} & \textbf{5.73}\\ \midrule
\multirow{9}{*}{GAT}          &    \multicolumn{0}{c|}{Retrain}                         & 82.26\(\pm\)0.35 & 12.55 & 74.18\(\pm\)0.46 & 19.31 & 85.14\(\pm\)0.22 & 77.42 & 88.63\(\pm\)0.56 & 48.05 & 82.04\(\pm\)0.38 & 103.11 & 65.99\(\pm\)0.13& 408.02 \\
                              &    \multicolumn{0}{c|}{GE-BLPA}                        & 69.13\(\pm\)0.02 & 8.66 & 68.42\(\pm\)0.03 & 11.01 & 74.92\(\pm\)0.00 & 30.46 & 77.19\(\pm\)0.01 & 18.60 & 73.35\(\pm\)0.03 & 26.12 & 59.16\(\pm\)0.07 & 56.56\\
                              &    \multicolumn{0}{c|}{GE-BKEM}                        & 52.92\(\pm\)0.00 & 7.18 & 53.53\(\pm\)0.00& 10.32 & 66.81\(\pm\)0.01 & 28.76 & 63.07\(\pm\)0.01 & 17.68 & 69.21\(\pm\)0.00& 24.86 & 56.97\(\pm\)0.05 & 57.92 \\
                              &    \multicolumn{0}{c|}{GUIDE-Fast}                     & 72.46\(\pm\)0.03 & 9.11 & 70.83\(\pm\)0.00 & 8.16 & 76.51\(\pm\)0.05 & 24.25 & 80.15\(\pm\)0.03 & 11.96 & 73.59\(\pm\)0.04 & 16.79 & OOM & OOM\\
                              &    \multicolumn{0}{c|}{GUIDE-SR}                       & 73.91\(\pm\)0.01 & 9.29 & 71.06\(\pm\)0.02 & 8.93 & 76.83\(\pm\)0.01 & 26.73 & 78.71\(\pm\)0.00 & 13.35 & 76.16\(\pm\)0.01& 18.43 & OOM & OOM\\
                              &    \multicolumn{0}{c|}{GNNDelete}                       & 76.87\(\pm\)0.11 & 1.94 & 72.18\(\pm\)0.20 & 2.21 & 81.25\(\pm\)0.12 & 3.77 & 85.10\(\pm\)0.25 & 2.47 & 79.15\(\pm\)0.17& 3.98 & 63.83\(\pm\)0.14 & 13.91  \\
                              &    \multicolumn{0}{c|}{GIF}                      & 77.44\(\pm\)0.05 & \textbf{0.53} & 71.30\(\pm\)0.13 & \textbf{0.61} & 78.68\(\pm\)0.06 & \textbf{1.23} & 85.77\(\pm\)0.08 & \textbf{0.91} & 79.08\(\pm\)0.19 & \textbf{1.12}  & OOM & OOM   \\
                              &    \multicolumn{0}{c|}{MEGU}                     & \underline{80.12\(\pm\)0.09} & \underline{0.86} & \underline{74.64\(\pm\)0.10} & \underline{1.02} & \textbf{83.26\(\pm\)0.14} & \underline{1.69} & \underline{87.91\(\pm\)0.12} & \underline{1.33} & \underline{81.35\(\pm\)0.07} & \underline{2.15} & \textbf{65.31\(\pm\)0.27} & \underline{10.56} \\
                              &    \multicolumn{0}{c|}{TCGU}                       & \textbf{82.36\(\pm\)0.37} & 3.34 & \textbf{76.14\(\pm\)0.24} & 3.68 & \underline{81.60\(\pm\)0.27} & 5.78 & \textbf{88.89\(\pm\)0.55} & 2.91 & \textbf{83.38\(\pm\)0.31}& 3.12 & \underline{65.10\(\pm\)0.94} & \textbf{7.30} \\ \bottomrule
\end{tabular}}
\end{table*}

\noindent \textbf{Datasets.} 
We evaluate our method on the node classification task with 6 benchmark datasets, including Cora, Citeseer, Pubmed~\cite{yang2016revisiting}, Amazon Photo, Amazon Computers~\cite{shchur2018pitfalls}, and a large graph Ogbn-Arxiv~\cite{hu2020open}. For the small and medium datasets, we randomly split nodes into 70\% for training, 10\% for validation, and 20\% for testing. For Ogbn-Arxiv, we adopt the public splits. Detailed Dataset descriptions and statistics can be found in Appendix~\ref{data_detail}.

\noindent \textbf{Baselines.} We compare TCGU to SOTA graph unlearning models that include 2 shard-based exact unlearning methods (GraphEraser~\cite{chen2022graph}, GUIDE~\cite{wang2023inductive}), and 3 general approximate unlearning methods (GNNDelete~\cite{cheng2022gnndelete}, GIF~\cite{wu2023gif}, and MEGU~\cite{li2024towards}). Besides, we add retraining from scratch (which is referred to as Retrain) as a natural competitor. For GraphEraser, we implement both balanced label propagation-based and embedding-based graph partition methods in the original paper and refer them to GE-BLPA and GE-BKEM, respectively. For GUIDE, we implement both fast partition (GUIDE-Fast) and spectral rotation-based partition (GUIDE-SR) in evaluation. For the 3 advanced approximate methods that are not applicable in immediate deletion scenarios, we keep the deleted data for them during the unlearning process.

\noindent \textbf{Evaluation.} In our experiments, we focus on three types of graph unlearning tasks. For node unlearning, following~\cite{li2024towards}, we randomly delete \(20\%\) of nodes in \(\mathcal{V}_{train}\) and their associated edges. For edge unlearning, we randomly delete edges in the training graph with an unlearning ratio of \(\rho\)\%. For feature unlearning, we randomly select \(\rho\)\% of nodes from \(\mathcal{V}_{train}\) and mask the full-dimensional features. Similar to~\cite{li2024towards}, the \(\rho\) is tuned in \([2.5,5,10,25,50]\). As for GNN backbones, following settings in~\cite{wu2023gif,li2024towards}, we use the two-layer GNN with 256 hidden units and train each model for 100 epochs before unlearning. Following~\cite{chen2022graph,li2024towards} we use the Micro-F1 score to evaluate semi-supervised node classification. We repeat each experiment 10 times for a fair comparison and report the average performance with standard deviation. We tailor the unlearning epochs for each GU strategy to its respective optimal values, ensuring convergence, and then record the average running time (in seconds). Additionally, to evaluate the unlearning efficacy (i.e., forgetting power) of TCGU, we adopt two attack strategies: Membership Inference Attack (MIA)~\cite{chen2021machine,chen2022graph} and Edge Attack~\cite{wu2023gif}. We give detailed descriptions of these two evaluation methods in Appendix~\ref{eval_attck}. 

\noindent \textbf{Hyperparameter settings.} In the condensation stage, the condensation ratio \(r_{cond}\) is tuned in a range of \(\{2.5\%,5\%,7.5\%,10\%\}\). We adopt a 3-layer MLP with 128 hidden units as topology function \(g_{\Phi}\). We further tune \(\lambda_{c}\) in a range of \(\{0.005,0.01,0.05,0.1\}\) and change \(\lambda_{f}\) from \(\{50,100,150,200\}\). For the fine-tuning stage, the rank \(r\) is searched in \(\{1,2,4,8\}\). We further tune \(T_{ft}\) in a range of \(\{15,20,25,30\}\). The trajectory length \(T_{s}\) is tuned amongst \(\{40,50,60\}\). More hyperparameter settings and implementation details can be found in Appendix~\ref{implement_details}.

\begin{table}[th]
\centering
\caption{Computational costs (s) of the TCGU pipeline.} 
\label{pipcost}
\renewcommand{\arraystretch}{0.9}
\resizebox{\linewidth}{!}{\begin{tabular}{c|cc|cc|cc}
\toprule
\multirow{2}{*}{Dataset} & \multicolumn{2}{c|}{Stage I} & \multicolumn{2}{c|}{Stage II} &  \multicolumn{2}{c}{Stage III} \\ 
              &   GCN        &  GAT        &      GCN        &  GAT  &  GCN        &  GAT       \\ \midrule
Cora                   & 44.0          &  48.5        &    1.76         & 2.41  & 0.38 &  0.93     \\ 
Citeseer                 & 72.3          &  82.2        &     1.85        &  2.81 &  0.34 &  0.87     \\
Pubmed                    & 132.7          &  210.5        &     3.66        &  4.96 & 0.74 & 0.82      \\
Photo                  &  106.3         &   91.9       &     1.70        &  2.02  &  0.25 &  0.89     \\
Computers                  & 152.5          & 253.2       &     2.54        &  2.59 & 0.22 & 0.53      \\
Ogbn-Arxiv                   &   326.6      &   398.0       &   4.85          &  6.14 & 0.88 &   1.16   \\\bottomrule
\end{tabular}}
\end{table}

\subsection{Evaluation of Model Utility}

To compare the predictive ability of models unlearned by different GU strategies, we report evaluation results of model utility in Table~\ref{utility}, which validates TCGU's superiority on most datasets.
For instance, it outperforms the best exact unlearning method GUIDE-SR by 8.45\% and 7.22\% in terms of F1-score with GAT on Cora and Computers, respectively. Besides, it exhibits an improvement of 2.14\% and 1.27\% than the most competitive approximate method MEGU with GCN trained on Citeseer and Photo, respectively. 
This validates TCGU can achieve better results even in the zero-glance setting, i.e., with less information. {Note that, the approximate methods need to access deleted targets during unlearning, which may be impractical in real-world applications.} In addition, TCGU achieves
a comparable performance with Retrain on all datasets and even outperforms it on half of the datasets (i.e. Cora, Citeseer, and Computers).
For a more comprehensive utility evaluation, we report results on more GNN backbones (SGC and GIN) and experimental analysis on different unlearning tasks (i.e., edge unlearning and feature unlearning) in Appendix~\ref{supply}.


\subsection{Evaluation of Unlearning Efficiency}

We report the average running time of different GU strategies 
in Table~\ref{utility}. It can be observed that TCGU significantly enhances unlearning efficiency compared with shard-based exact methods. For example, with GAT trained on small datasets Cora and Citeseer, TCGU is at least 2.2 times faster than four exact unlearning methods. On larger datasets Computers and Arxiv, TCGU can shorten the running time by about 4.7\(\sim\)8.8 times. Additionally, TCGU can achieve comparable efficiency to 3 approximate unlearning methods on all datasets. The efficiency gap becomes even smaller as the graph gets larger. On large-scale Arxiv, it takes only 5.73s for TCGU to unlearn the GCN model, which is the most efficient of all methods. This is because data transfer and model retraining on condensed data are highly efficient.

\noindent \textbf{Time Cost Analysis} To further investigate the time cost of our pipeline, we report averaged running time in each of the three stages with GCN and GAT on 6 datasets in Table~\ref{pipcost}. Pre-condensation (Stage I) requires significantly higher computation than the unlearning process which comprises the data transfer(Stage II) and the retraining (Stage III). However, since pre-condensation is only performed once and enables the use of condensed data for subsequent unlearning requests (see Section~\ref{seq_unlearn}), we can tolerate this
cost.
The two stages during unlearning are highly efficient. For Stage II, our fine-tuning technique avoids re-condensing data from scratch, leveraging the knowledge gained during the initial condensation. The low-rank plugin further reduces computation in data transfer and stabilizes training. Additionally, retraining the GNN on condensed graphs takes less than one second, as the small dataset size makes training highly efficient. A detailed time complexity analysis for each stage is provided in Appendix~\ref{complexity}.

\subsection{Evaluation of Unlearning Efficacy}
\label{unlearn_efficacy}

\begin{table}[t]
\caption{Comparasion of unlearning efficacy with AUC of membership inference attack}
\label{mia}
\centering
\resizebox{\linewidth}{!}{\begin{tabular}{ccccccccc}
\toprule
\multirow{2}{*}{Method} & \multicolumn{2}{c}{Cora} &  & \multicolumn{2}{c}{Photo} &  & \multicolumn{2}{c}{Arxiv} \\ \cline{2-3} \cline{5-6} \cline{8-9} 
                  & GCN      & GAT          &  &     GCN    & GAT        &  &      GCN    & GAT       \\ \midrule
Retrain           &         0.502 &  0.509          &  &    0.510& 0.515         &  & 0.504          &  0.503        \\
GNNDelete                  &      \underline{0.559} &  \underline{0.556}      &   &         \underline{0.541} & 0.594         &  &   0.553        &  0.572      \\
GIF                  &      0.586 &  0.671      &  &      0.579& 0.602         &  & OOM          & OOM         \\ 
MEGU                  &    0.565&  0.548   &  &      0.554 & \underline{0.575}         &  &     \underline{0.529}      &  \underline{0.547}         \\ 
TCGU                  &    \textbf{0.524}&  \textbf{0.527}     &  &         \textbf{0.514} & \textbf{0.508}        &  &    \textbf{0.521}      &    \textbf{0.516}      \\ \bottomrule
\end{tabular}}
\end{table}

\begin{figure}[!h]
	\centering
	\subfloat[GCN]{
		\includegraphics[width=0.49\linewidth]{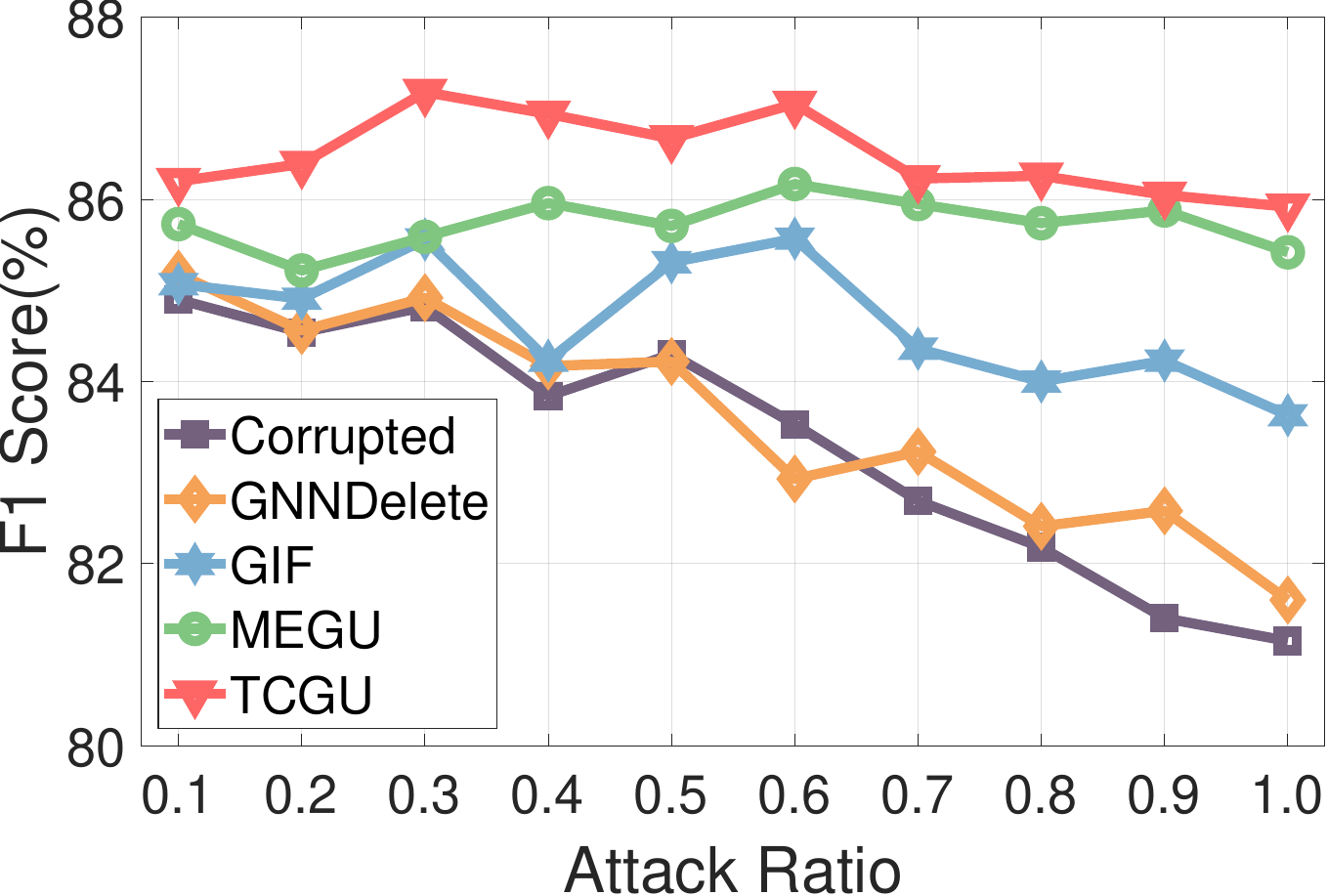}}
	\subfloat[GAT]{
		\includegraphics[width=0.49\linewidth]{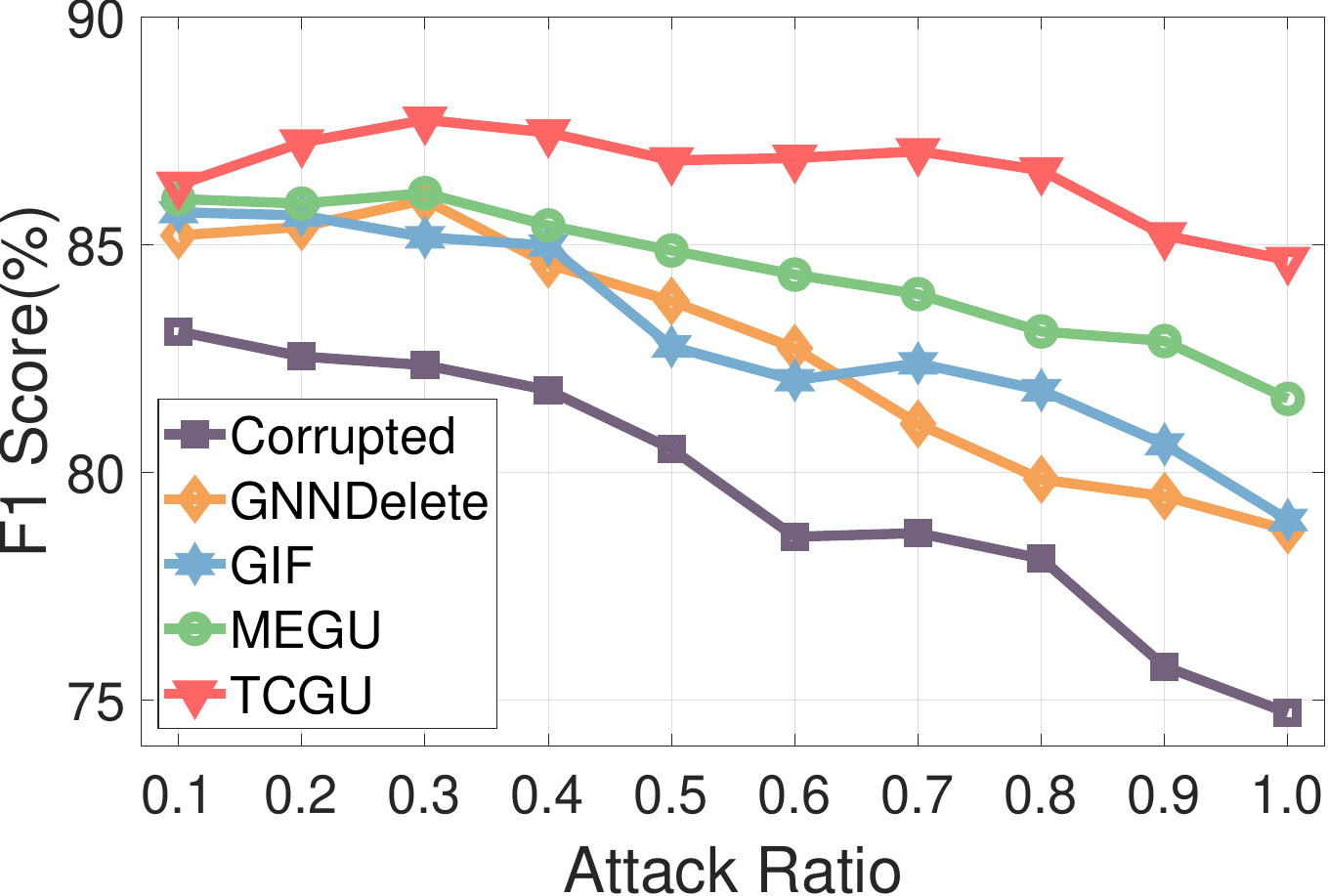}}
        \caption{Comparasion of unlearning efficacy with adversarial edge attack}
        \label{fig:edgeattack}
\end{figure}

To measure the degree of data removal (unlearning efficacy), we adopt two commonly used attack-based methods: Membership Inference Attack (MIA) and Edge Attack (EA). 

\noindent \textbf{Membership Inference Attack.} Following the settings in~\cite{chen2022graph}, we apply MIA to quantify extra information leakage after node unlearning. Higher attack AUC indicates more target unlearned samples would be predicted by attack models. As shown in Table~\ref{mia}, we observe that attack AUC on TCGU remains closer to 0.5 (random guess) than 3 advanced approximate methods on 3 datasets, demonstrating its superior privacy-preserving capability. In further analysis, on the Cora and Photo datasets, the attack AUC for GIF consistently exceeds 0.564, even surpassing 0.67 with GAT trained on Cora. We also observe that the attack AUC on MEGU is above 0.57 on both Photo and Arxiv with GAT model. These results show the privacy leakage concern in existing approximate unlearning approaches. In contrast, our TCGU reduces the attack AUC to no more than 0.527 in all 6 cases (3 datasets and 2 GNN backbones for each of them). This validates that our data-centric method can effectively erase the influence of deleted data.

\noindent \textbf{Edge Attack.} Following~\cite{li2024towards,wu2023gif}, we perform an adversarial edge attack on Cora to mislead graph representation learning with different noise ratios. The adversarial edges are treated as unlearning targets in this experiment. Therefore, a larger utility gain indicates higher unlearning efficacy. From Figure~\ref{fig:edgeattack}, we find GNN models suffer a performance drop training on the corrupted graphs as the attack ratio increases. These results show that
adversarial edges can greatly impair model utility. While GNNDelete and GIF show significant downward trends as the noise ratio increases, TCGU is better at preserving model utility and consistently achieves an F1-score above 0.84 with various backbones, even at an attack ratio of 1.0. Compared with competitive MEGU, TCGU also achieves a more robust model performance. For example, on the GAT model, when the attack ratio is increased from 0.5 to 1.0, the relative F1-score improvement is about 1.97\% to 3.54\%. The above results demonstrate that TCGU can effectively unlearn the adverse impacts caused by noise edges and achieve better model robustness than existing approximate GU strategies.

\subsection{Ablation Analysis}

We perform ablation studies on Cora, Citeseer, and Computers to verify the effectiveness of well-designed techniques in TCGU, including 2 alignment objectives in the pre-condensation (i.e., CFA and LA) and 4 strategies used in the data transfer stage (i.e., LoRP, TFS, SDM, and CDR).
For TCGU w/o TFS, we replace TFS with random network sampling~\cite{yuan2023real,zheng2023rdm}. 

The results are illustrated in Table~\ref{ablation}. In general, the full TCGU achieves a better trade-off between model utility and forgetting power than different variants. Specifically, the model performance significantly drops when removing FA or LA. This validates the effectiveness of our two-level alignment condensation strategy. We also notice that these two variants even have lower attack AUC than full TCGU after data transfer. The reason is that the low-quality pre-condensation cannot effectively capture the original graph knowledge, which makes synthetic data less meaningful. Note that there is an evident increasing attack AUC when transferring data without the low-rank plugin and the model performance remains high. This shows that without the proposed residual plugin, directly fine-tuning pre-condensed data cannot effectively erase the influence of deleted data. Besides, we find TCGU achieves a performance gain when equipped with trajectory-based function sampling. This is because more diverse and pattern-preserving embedding functions can help MMD computing in DM. For TCGU-w/o SDM, we find there is not only a utility drop, but an increase in privacy concerns. For instance, On Cora with GCN, the attack AUC is 9.0\% higher (0.524 \(\rightarrow\) 0.571) without SDM. This is due to that the data cannot be effectively transferred into the new distribution and causes information leakage. 
The ablation results of the above three techniques in data transfer also validate that the model trained on the pre-condensed data still suffers from privacy issues as the small-scale data contains training knowledge on the original data. The unlearning power of TCGU is mainly attributed to our fine-tuning operation. We also find the contrastive regularizer is highly effective in utility-preserving. For example, with GAT as the backbone, TCGU outperforms the TCGU w/o CDR by 5.46\% and 7.96\% inF1-score on the Citeseer and Computers, respectively.

\begin{table}
\caption{Ablation study}
\label{ablation}
\centering
\resizebox{\linewidth}{!}{\begin{tabular}{cc|cccccc}
\toprule
\multirow{2}{*}{GNN} & \multicolumn{1}{c|}{\multirow{2}{*}{Component}} & \multicolumn{2}{c}{Cora} & \multicolumn{2}{c}{Citeseer} & \multicolumn{2}{c}{Computers} \\
                     & \multicolumn{1}{c|}{}                        & F1        & AUC  & F1        & AUC & F1        & AUC      \\ \midrule
\multirow{7}{*}{GCN} &    w/o FA                                    &  44.36\(\pm\)0.27         &    0.523   &   35.64\(\pm\)0.47        & 0.519 & 55.05\(\pm\)0.39  &  \underline{0.517}    \\
                     &    w/o LA                                    & 62.11\(\pm\)0.40         &    \underline{0.519}   &    69.92\(\pm\)0.31       & \textbf{0.515} &     58.47\(\pm\)0.34      &  \textbf{0.514}     \\
                    &    w/o LoRP                                    &    \textbf{84.54\(\pm\)0.33}     &   0.588    & \underline{75.34\(\pm\)0.33}        & 0.552 &    \underline{84.35\(\pm\)0.28}      & 0.573      \\
                     &    w/o TFS                                         &    77.82\(\pm\)0.29       &  0.531  &  73.39\(\pm\)0.17         & 0.538 &  81.05\(\pm\)0.20       & 0.529         \\
                     &    w/o SDM                                          &  76.52\(\pm\)0.31         &  0.571  &  72.18\(\pm\)0.42         & 0.560 &     82.24\(\pm\)0.46      &  0.545           \\
                     &    w/o CDR                                           &   78.19\(\pm\)0.44        &   \textbf{0.512}  &     72.56\(\pm\)0.35      & 0.522 &   81.31\(\pm\)0.37       & 0.521              \\
                     &    TCGU                                          &   \underline{82.18\(\pm\)0.26}        & 0.524   &    \textbf{75.99\(\pm\)0.18}         & \underline{0.516} &     \textbf{84.76\(\pm\)0.22}      &     0.519        \\ \midrule
\multirow{7}{*}{GAT} &    w/o FA                                   &    33.09\(\pm\)0.31       &  \underline{0.518}  &    27.67\(\pm\)0.52       & \underline{0.509} &   37.96\(\pm\)0.59        &    \underline{0.503}          \\
                     &    w/o LA                                   & 62.48\(\pm\)0.46          &  \textbf{0.509}  &  50.98\(\pm\)0.43         & 0.519 &    58.15\(\pm\)0.30      &   0.507           \\
                    &    w/o LoRP                                    &   \textbf{84.10\(\pm\)0.36}      &  0.563    &    \underline{73.23\(\pm\)0.34}     & 0.532 &    \underline{81.05\(\pm\)0.28}      & 0.556       \\
                     &    w/o TFS      &  79.67\(\pm\)0.33        &  0.535  &    72.63\(\pm\)0.25        & 0.522 &   80.41\(\pm\)0.19        &   0.517           \\
                     &    w/o SDM      &  75.01\(\pm\)0.25         &  0.542  &   72.78\(\pm\)0.32        & 0.544 &   79.08\(\pm\)0.40        &    0.528          \\
                     &    w/o CDR       &  79.48\(\pm\)0.38         & 0.528   &    70.68\(\pm\)0.27       & 0.511 &  75.42\(\pm\)0.25         &   \textbf{0.501}           \\
                     &    TCGU         &     \underline{82.36\(\pm\)0.37}      & 0.527   &   \textbf{76.14\(\pm\)0.24}        & \textbf{0.507} &    \textbf{83.38\(\pm\)0.31}       & 0.514             \\ \bottomrule
\end{tabular}}
\end{table}

\subsection{Sequential Unlearning Evaluation}
\label{seq_unlearn}

\begin{figure}[t]
	\centering
	\subfloat[Cora (Utility)]{
		\includegraphics[width=0.45\linewidth]{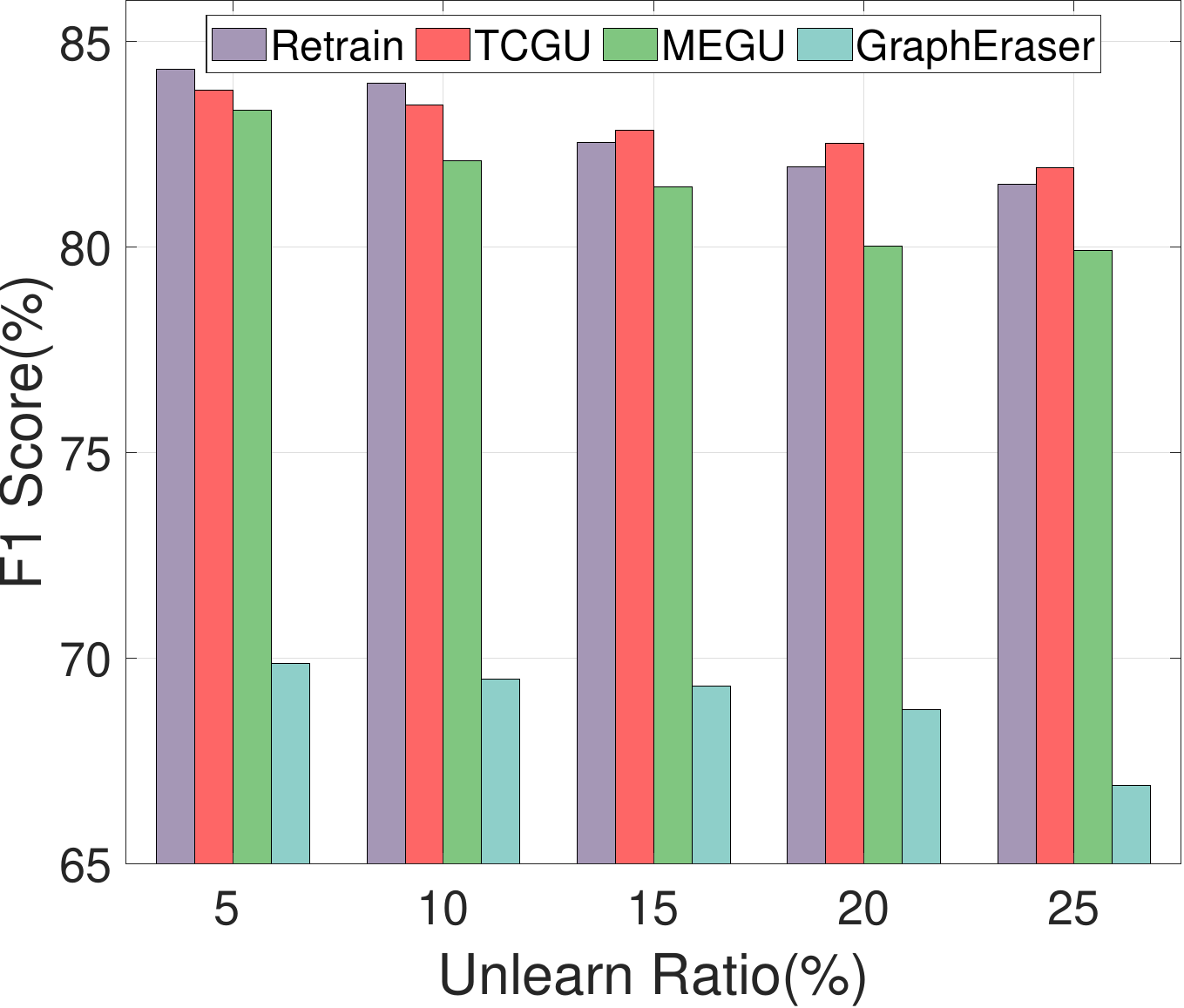}}
	\subfloat[Cora (Attack)]{
		\includegraphics[width=0.45\linewidth]{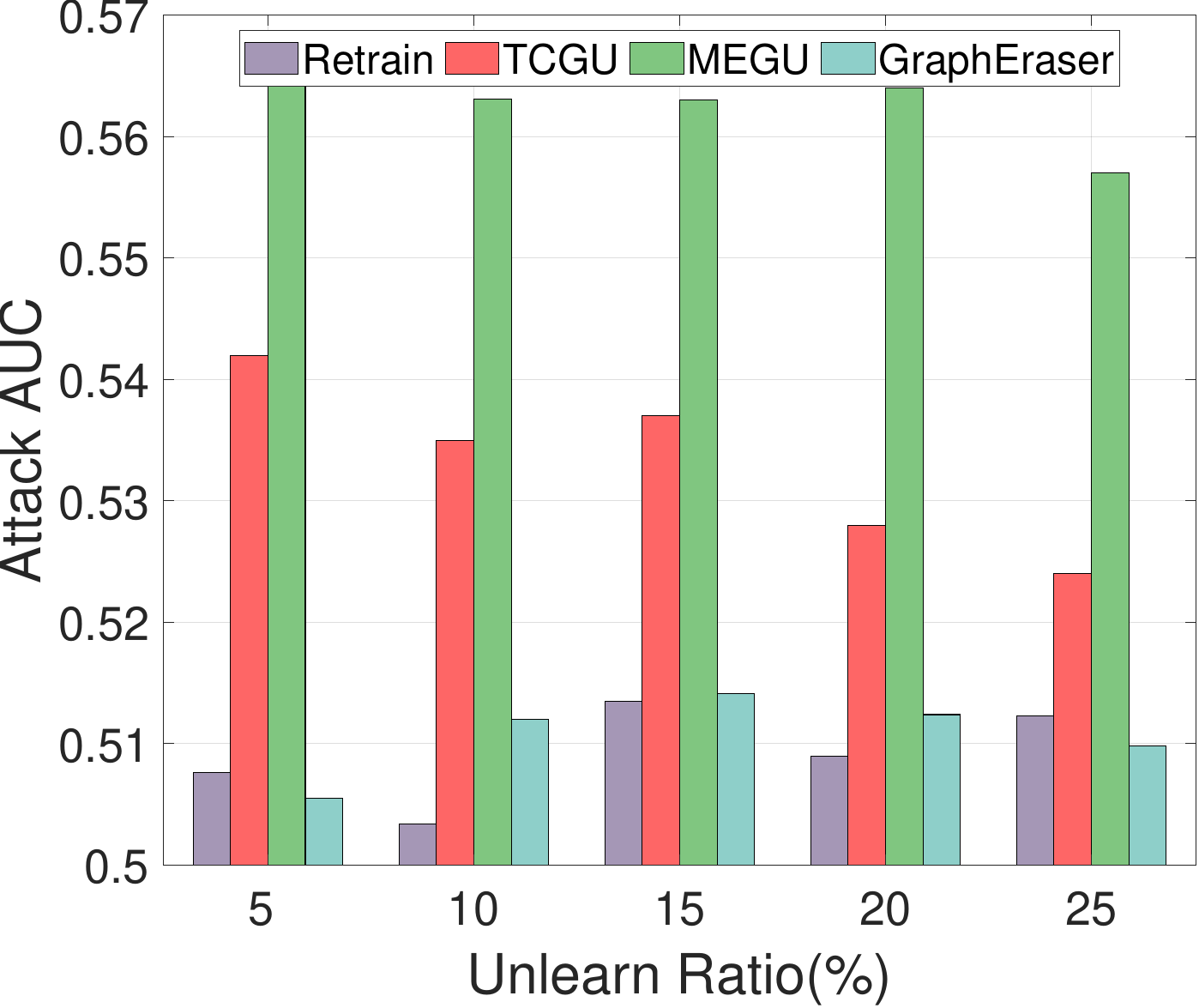}}
        \\
        \subfloat[Computers (Utility)]{
		\includegraphics[width=0.45\linewidth]{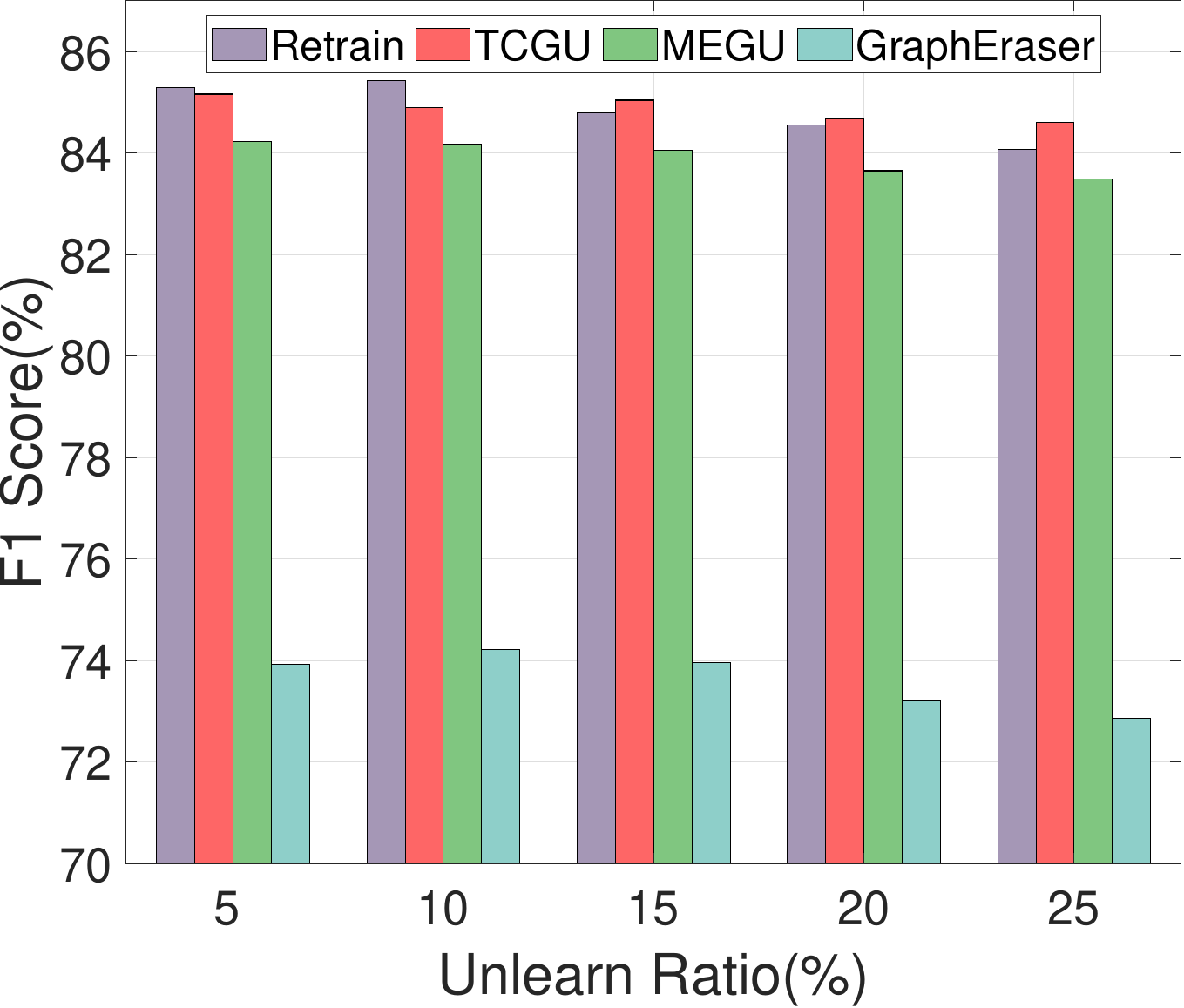}}
        \subfloat[Computers (Attack)]{
		\includegraphics[width=0.45\linewidth]{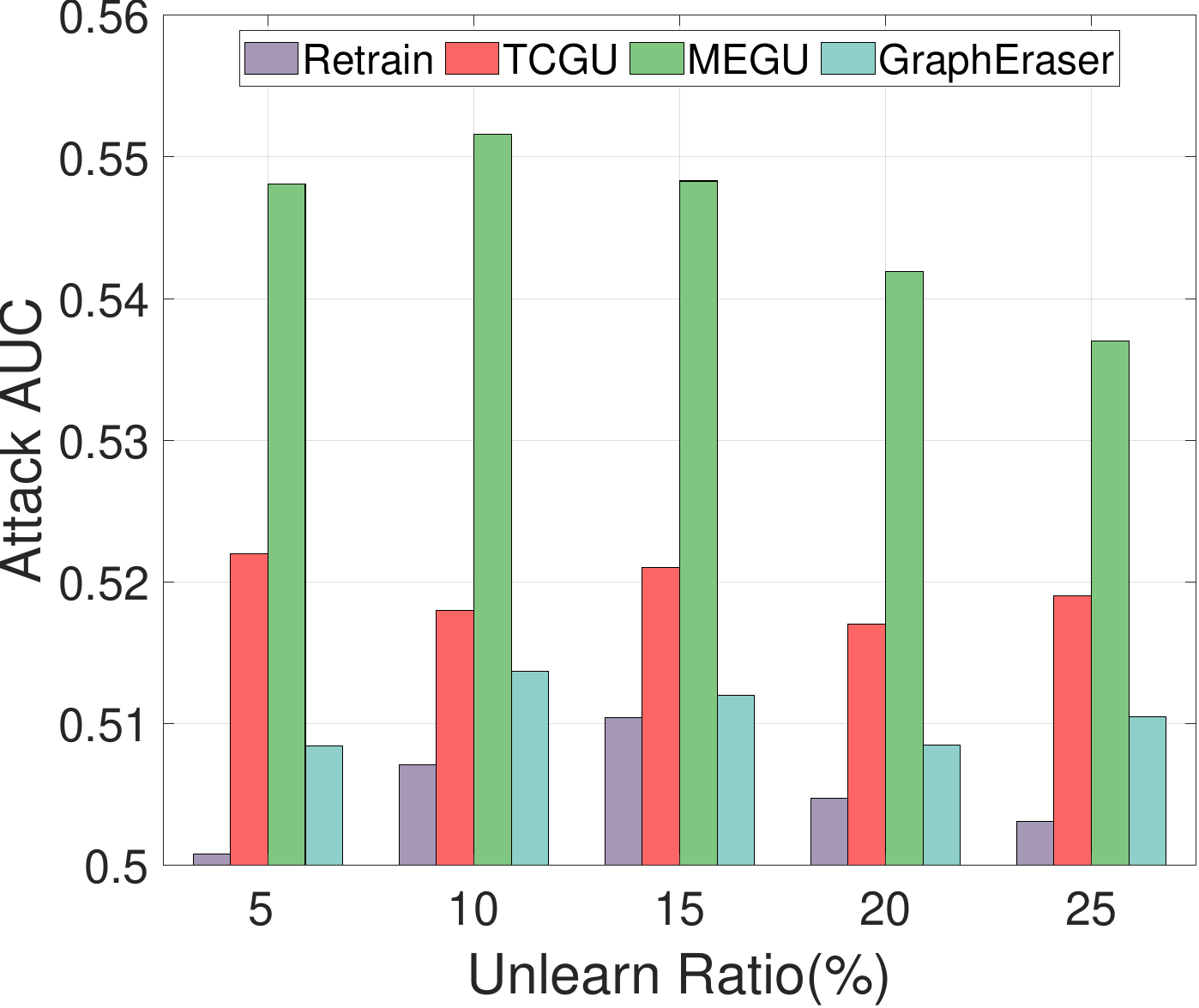}}
        \caption{Sequential Unlearning Experiments}
        \label{fig:continulearn}
\end{figure}

TCGU is designed to handle a sequence of deletion requests. After pre-condensation, it supports sequential unlearning by continuing to fine-tune the condensed data as new unlearning requests arrive. To validate this ability, we conduct a sequential deletion of 25\% of nodes with a batch size of 5\% on the Cora and Computers datasets. The utility and unlearning efficacy of different GU methods during sequential unlearning are shown in Fig~\ref{fig:continulearn}. We observe that TCGU consistently achieves superior utility and unlearning efficacy than the SOTA approximate method MEGU across different deletion ratios. Moreover, as deletion requests increase, the performance drop in utility for TCGU is smaller than that of the Retrain method. Notably, when the unlearning ratio exceeds 15\%, TCGU achieves the best prediction performance on both datasets. Additionally, TCGU maintains stable unlearning efficacy during sequential unlearning, with the attack AUC on Computers consistently below 0.522 and the AUC on Cora decreasing from over 0.54 to 0.524. These results confirm TCGU's effectiveness in sequential unlearning.

\section{Conclusion}

In this work, we investigate the graph unlearning problem in the zero-glance setting and propose TCGU, a novel data-centric solution based on transferable data condensation.
We first distill the original graph into a small and utility-preserving proxy dataset with a two-level alignment condensation strategy. Then, to perform unlearning with condensed data, we introduce a low-rank-based fine-tuning technique to align pre-condensed data with the remaining graph efficiently. The novel similarity distribution matching objective together with covariance-preserving feature alignment can effectively transfer intra- and inter-class relationships of remaining data to the updated condensed graph.
A contrastive-based regularizer is also developed to make features extracted from condensed data more discriminative, thus facilitating GNN training and preserving model utility.
The unlearned model is obtained by retraining on the transferred condensed data. 
Extensive experiments on 6 benchmark datasets confirm the superiority of our TCGU in terms of model utility, unlearning efficiency, and unlearning
efficacy.

\bibliographystyle{ACM-Reference-Format}
\bibliography{sample-base.bib}

\newpage
\appendix

\section{Detailed Experimental Settings}
\label{expset}

\subsection{Dataset Details}
\label{data_detail}

\begin{table}[h]
\caption{Statistics of the experimental datasets}
\label{tab:stat}
\centering
\resizebox{\linewidth}{!}{\begin{tabular}{ccccc}
\toprule
Dataset & \#Nodes & \#Edges & \#Features & \#Classes  \\ \midrule
Cora & 2,708 & 5,429 &  1,433 & 7       \\
Citeseer & 3,327 & 4,732 & 3,703 & 6    \\ 
Pubmed  & 19,717 & 44,338 & 500 & 3       \\ 
Amazon Photo & 7,487 & 119,043 & 745 & 8    \\ 
Amazon Computers  & 13,381 & 245,778 & 767 & 10  \\ 
Ogbn-Arxiv  & 169,343 & 1,166,243 & 128 & 40  \\\bottomrule
\end{tabular}}
\end{table}

\noindent\textbf{Cora}, \textbf{Citeseer}, and \textbf{Pubmed} are three citation networks where the nodes represent the publications, and edges represent citation relationships. The node features are binary
vectors indicating the presence of the keywords from a dictionary,
and the class labels represent the publications’ research field.

\noindent\textbf{Amazon Photo} and \textbf{Amazon Computers} are segments of the Amazon co-purchase graph~\cite{mcauley2015image}, where nodes represent goods, edges indicate that two goods are frequently bought together, node features are bag-of-words encoded product reviews, and class labels represent product category.

\noindent\textbf{Ogbn-Arxiv} is a citation network between all Computer Science (CS) ARXIV papers indexed by MAG~\cite{wang2019deep}. Each node is an ARXIV
paper and each edge indicates the citation relationship. Each paper comes with a 128-dimensional feature vector obtained by averaging the embeddings of words in its title and abstract.

The statistics of the above datasets are summarized in Table~\ref{tab:stat}

\subsection{Evaluation Method for Unlearning Efficacy}
\label{eval_attck}

\textbf{Membership Inference Attack.} Following~\cite{chen2022graph,wang2023inductive}, we apply an enhanced membership inference attack~\cite{chen2021machine}, the attacker with access to both the original and the unlearned model could determine whether a specific node is indeed revoked from the unlearned model. 
We report the AUC of the membership inference attack on the model. The higher value indicates more extra information leakage.

\noindent \textbf{Advesarial Edge Attack.} Following~\cite{li2024towards,wu2023gif}, we randomly add a certain number of edges to the training graph, satisfying that two nodes associated with the newly added edge are from different classes. These noise edges will mislead the representation learning, thus impairing the prediction performance. Thereafter, we treat them as unlearning targets and assess the utility of the unlearned model. Intuitively, if a method achieves better unlearning efficacy,
it tends to effectively mitigate the negative impact of noisy
edges on predictive performance, thus ensuring robustness.

\subsection{Hyperparameters \& Implementation}
\label{implement_details}

The sparsification threshold \(\delta\) is set to 0.05 for 5 datasets. We fix self-loop weight \(w_{loop}\) to 1 and hop number \(K\) to 2 for all datasets. 
We tune the condensation epoch \(T_{cond}\) in a range of \(\{1500, 2000,2500\}\). 
We change \(\tau_{s}\) within \(\{2,5,10\}\) and \(L_{s}\) in \(\{5,10,20\}\). The queue volume \(N_{\mathcal{Q}}\) is set to 20. 
We tune the regularization coefficient \(\lambda_{r}\) in a range of \(\{5e^{-4},1e^{-3},2e^{-3},5e^{-3}\}\).
We implement TCGU with Pytorch 1.12~\cite{paszke2019pytorch} and Python 3.9. All experiments are conducted on a Linux server with an Intel Xeon Silver 4208 CPU, a Quadro RTX 6000 24GB GPU, and 128GB of RAM. 


\section{Algorithm}
\label{method_supp}

\renewcommand{\algorithmicrequire}{\textbf{Input:}}
\renewcommand{\algorithmicensure}{\textbf{Output:}}

\begin{algorithm}
\caption{Tow-level Alignment Graph Condensation (TAGC)}\label{alg:gc}
\begin{algorithmic}[1]
\Require \(\mathcal{G}(\mathbf{A},\mathbf{X},\mathbf{Y})\): original graph; \(\textbf{GNN}_{\theta_{\mathcal{G}}}\): pre-trained model; \(\mathbf{Y}^{\prime}\): pre-defined condensed labels; \(K\): hop number; \(T_{cond}\): pre-condensation step;
\Ensure Condensed graph \(\mathcal{G}^{\prime}(\mathbf{A}^{\prime},\mathbf{X}^{\prime},\mathbf{Y}^{\prime})\)
\State Initialize \(\mathbf{X}^{\prime}\) as node features randomly selected from each class
\State Initialize \(g_{\Phi}\) as the topology function randomly with an MLP.
\State Compute \(K\)-hop node features \(\{\mathbf{H}^{(k)}\}_{k=0}^{K}\) for \(\mathcal{G}\)
\State Compute \(\mu_{c}^{(k)}\) and  \(U_{c}^{(k)}\), with \(k=0,1,...K\) and \(c=1,2,...C\)
\For{$t=0,...,T_{cond}-1$}
\State Compute \(\mathbf{A}^{\prime}=g_{\Phi}(\mathbf{X}^{\prime})\)
\State Compute \(K\)-hop node features \(\{\mathbf{H}^{\prime (k)}\}_{k=0}^{K}\) for \(\mathcal{G}^{\prime}\)
\State \(\mathcal{L}_{mean},\mathcal{L}_{cov} \leftarrow (0,0)\)
\For{$c=1,...,C$}
\For{$k=0,...,K$}
\State Compute \(\mu_{c}^{\prime (k)}\) and \(U_{c}^{\prime (k)}\)
\State \(\mathcal{L}_{mean} \leftarrow \mathcal{L}_{mean}+r_{c} \cdot \Vert \mu_{c}^{(k)}-\mu_{c}^{\prime (k)} \Vert_{2}^{2}\)
\State \(\mathcal{L}_{cov} \leftarrow \mathcal{L}_{cov}+r_{c} \cdot \Vert U_{c}^{(k)}-U_{c}^{\prime (k)} \Vert_{2}^{2}\)
\EndFor
\EndFor
\State \(\mathcal{L}_{feat}=\mathcal{L}_{mean}+\lambda_{c}\mathcal{L}_{cov}\) 
\State \(\mathcal{L}_{logits} = \mathcal{L}_{cls}(\text{GNN}_{\theta_{\mathcal{G}}}(\mathbf{A}^{\prime},\mathbf{X}^{\prime}),\mathbf{Y}^{\prime})\) 
\State \(\mathcal{L}_{cond} = \mathcal{L}_{logits} + \lambda_{f}\mathcal{L}_{feat}\)
\If{$t\%(\tau_{1}+\tau_{2}) \textless \tau_{1}$} 
\State Update \(\mathbf{X}^{\prime} \leftarrow \mathbf{X}^{\prime}-\eta_{1}\nabla_{\mathbf{X}^{\prime}}\mathcal{L}_{cond}\)
\Else 
\State Update \(\Phi \leftarrow \Phi-\eta_{2}\nabla_{\Phi}\mathcal{L}_{cond}\)
\EndIf
\EndFor
\State \(\mathbf{A}^{\prime}=\mathrm{RELU}(g_{\Phi}(\mathbf{X}^{\prime})-\delta)\)
\end{algorithmic}
\end{algorithm}

\begin{algorithm}
\caption{Efficient Fine-tuning for Condensed Data Transfer}\label{alg:ft}
\begin{algorithmic}[1]
\Require \(\mathcal{G}_{r}(\mathbf{A}_{r},\mathbf{X}_{r},\mathbf{Y}_{r})\): remaining graph; \(\mathbf{X}^{\prime}\): pre-condensed node features; \(g_{\Phi}\): topology function; \(\mathbf{Y}^{\prime}\): condensed labels; \(K\): hop number; \(r\): rank number; \(T_{ft}\): fine-tuning step; \(N_{\mathcal{Q}}\): function queue volume; \(T_{s}\): trajectory length; \(\tau_{s}\): sample interval; \(L_{s}\): function sample size;
\Ensure Unlearned condensed graph \(\mathcal{G}_{u}^{\prime}=(\mathbf{A}^{\prime}_{u},\mathbf{X}^{\prime}_{u},\mathbf{Y}^{\prime}_{u})\)
\State \(\mathcal{Q} \gets \emptyset, \mathbf{Y}^{\prime}_{u} \gets \mathbf{Y}^{\prime}\)
\State Initialize \(\mathcal{A},\mathcal{B}\) with a rank of \(r\)
\State Freeze pre-condensed node features \(\mathbf{X}^{\prime}\)
\State Compute \(K\)-hop node features \(\{\mathbf{H}^{(k)}_{r}\}_{k=0}^{K}\) for \(\mathcal{G}_{r}\)
\For{$t=0,...,T_{ft}-1$}
\State \(\mathbf{X}^{\prime}_{u}=\mathbf{X}^{\prime}+\mathcal{A}\mathcal{B}\) 
\State Compute \(\mathbf{A}^{\prime}_{u}=g_{\Phi}(\mathbf{X}^{\prime}_{u})\)
\If{$t\%\tau_{s}=0$} 
\State Randomly Initialize a GNN \(f_{\theta_{0}}\) with parameter \(\theta_{0}\)
\For{$t^{\prime}=0,...,T_{s}-1$}
\If{$t^{\prime}\%\lceil \frac{T_{s}}{L_{s}} \rceil=0$}
\State Push a new model \(f_{\theta_{t^{\prime}}}\) into \(\mathcal{Q}\)
\State Pop a model from \(\mathcal{Q}\) if \(\lvert \mathcal{Q} \rvert \textgreater N_{\mathcal{Q}}\)
\EndIf
\State \(\theta_{t^{\prime}+1} = \theta_{t^{\prime}}- \nabla_{\theta_{t^{\prime}}}\mathcal{L}_{cls}(f_{\theta_{t^{\prime}}}(\mathbf{A}^{\prime}_{u},\mathbf{X}^{\prime}_{u}),\mathbf{Y}^{\prime}_{u})\)
\EndFor
\EndIf
\State Sample embedding function \(f_{\theta}\) from \(\mathcal{Q}\) randomly
\State  \(Z_{r} = f_{\theta}(\mathbf{A}_{r},\mathbf{X}_{r}), Z^{\prime}_{u} = f_{\theta}(\mathbf{A}^{\prime}_{u},\mathbf{X}^{\prime}_{u})\)
\State Compute similarity embedding matrices \(S_{r}, S_{u}^{\prime}\) by Eq.~\ref{sim}
\State Compute \(\mathcal{L}_{sm}\) according to Eq.~\ref{sim_loss} 
\State Compute \(\mathcal{L}_{cdr}\) according to Eq.~\ref{cdr} 
\State Compute \(\mathcal{L}_{feat}\) according to Eq.~\ref{feat_loss} 
\State \(\mathcal{L}_{ft} = \mathcal{L}_{sm} + \lambda_{f}^{\prime}\mathcal{L}_{feat}+\lambda_{r}^{\prime}\mathcal{L}_{cdr}\)
\If{\(t\%(\tau_{1}^{\prime}+\tau_{2}^{\prime}) \textless \tau_{1}^{\prime}\)}
\State Update \(\mathcal{A} \leftarrow \mathcal{A}-\eta_{1}^{\prime}\nabla_{\mathcal{A}}\mathcal{L}_{ft}\) and \(\mathcal{B} \leftarrow \mathcal{B}-\eta_{1}^{\prime}\nabla_{\mathcal{B}}\mathcal{L}_{ft}\) 
\Else 
\State Update \(\Phi \leftarrow \Phi-\eta_{2}^{\prime}\nabla_{\Phi}\mathcal{L}_{ft}\)
\EndIf
\EndFor
\State \(\mathbf{A}^{\prime}_{u}=\mathrm{RELU}(g_{\Phi}(\mathbf{X}^{\prime}_{u})-\delta)\)
\end{algorithmic}
\end{algorithm}

\subsection{Overall Algorithm of TAGC}

We illustrate the overall algorithm of TAGC in Algorithm~\ref{alg:gc}. We first set the pre-defined condensed labels \(\mathbf{Y}^{\prime}\) and randomly initialize node features \(\mathbf{X}^{\prime}\) and function \(g_{\Phi}\). Then, we pre-compute multi-hop node features \(\{\mathbf{H}^{(k)}\}_{k=0}^{K}\) of \(\mathcal{G}\) and then calculate mean feature \(\mu_{c}^{(k)}\) and covariance matrix \(U_{c}^{(k)}\) for each hop and class. During condensation, we compute \(\{\mathbf{H}^{\prime (k)}\}_{k=0}^{K}\) for \(\mathcal{G}^{\prime}\) and align it with \(\{\mathbf{H}^{ (k)}\}_{k=0}^{K}\) in both mean and covariance aspects within each class to get feature alignment loss. After that, we calculate logits alignment loss with pre-trained GNN. The sum of losses is used to update \(\mathbf{X}^{\prime}\) and \(\Phi\). We optimize these two parts alternatively until reaching maximum epoch number \(T_{cond}\). Finally, we obtain the sparsified graph structure as \(\mathbf{A}^{\prime}=\mathrm{RELU}(g_{\Phi}(\mathbf{X}^{\prime})-\delta)\).

\subsection{Overall Algorithm of Data Transfer}

We show the detailed algorithm of condensed data transfer in Algorithm~\ref{alg:ft}. In detail, we first freeze parameters of the learned feature matrix \(\mathbf{X}^{\prime}\) and initialize low-rank plugins \(\mathcal{A},\mathcal{B}\) and function queue \(\mathcal{Q}\). During the data transfer stage, we add the multiplication of plugins behind frozen \(\mathbf{X}^{\prime}\) and get updated features \(\mathbf{X}^{\prime}_{u}\). Then, we perform trajectory-based function sampling every \(\tau_{s}\) fine-tuning steps and push \(L_{s}\) new models into \(\mathcal{Q}\) each time. After randomly sampling a function \(f_{\theta}\) from \(\mathcal{Q}\), we encode \(\mathcal{G}_{r}\) and \(\mathcal{G}_{u}^{\prime}\) into embedding space and calculate similarity distribution matching loss, regularization discrimination loss. Besides, we calculate feature alignment loss in the transfer stage. The weighted sum of these three losses is used to update \(\mathcal{A},\mathcal{B}\) and \(\Phi\) alternatively as TAGC. The optimization process finishes until the maximum fine-tuning step \(T_{ft}\). Similar to TAGC, we infer the final sparsified adjacency matrix \(\mathbf{A}^{\prime}_{u}=\mathrm{RELU}(g_{\Phi}(\mathbf{X}^{\prime}_{u})-\delta)\).

\section{Theoratical Analysis}
\label{theory}

\subsection{Proof of Propositions}

\noindent \textbf{Proof of Proposition 1.}
The proof of proposition 1 is derived by extending the proposition in~\cite{gao2024rethinking}.
\begin{proof}
    Let the propagated node features for the original graph $\mathcal{G}$ and the condensed graph $\mathcal{G^{\prime}}$ as:
\begin{equation}
    \mathbf{H} = \mathbf{P^{k}}\mathbf{X}, \quad \mathbf{H^{\prime}} = \mathbf{P^{\prime k}}\mathbf{X^{\prime}} 
\end{equation}
Without loss of generality, we use the gradient matching (GM) objective from GCond~\cite{jin2021graph} as an example, where SGC serves as the relay model. As a result, the prediction results can be expressed as:
\begin{equation}
        \mathbf{Z}=\mathbf{H}\mathbf{\Theta}, \quad \mathbf{Z^{\prime}}=\mathbf{H^{\prime}}\mathbf{\Theta},  
\end{equation}
where $\mathbf{\Theta}$ is the learnable parameters. As the GM objective is calculated class-wise, we simplify the objective of the model into the MSE loss and calculate it separately for each class:
\begin{equation}
    \begin{split}
        \mathcal{L}_{i}^{\mathcal{G}}&=\frac{1}{2}\|\mathbf{H_{i}}\mathbf{\Theta}-\mathbf{Y_{i}}\|^{2} \\
    \mathcal{L}_{i}^{\mathcal{G^{\prime}}}&=\frac{1}{2}\|\mathbf{H^{\prime}_{i}}\mathbf{\Theta}-\mathbf{Y^{\prime}_{i}}\|^{2}, \\
    \end{split}
\end{equation}
where $\mathbf{Y_{i}},Y^{\prime}_{i}$ are one-hot labels for nodes in class $i$ in $\mathcal{G}$ and $\mathcal{G^{\prime}}$, respectively. The GM objective can be formulated as:
\begin{equation}
    \begin{split}
    &\mathcal{L}_{GM}=\sum_{i=1}^{c}\bigg\|\frac{1}{|C_{i}|}\nabla_{\mathbf{\Theta}}\mathcal{L}_{i}^{\mathcal{G}}-\frac{1}{|C_{i}^{\prime}|}\nabla_{\mathbf{\Theta}}\mathcal{L}_{i}^{\mathcal{G}^{\prime}}\bigg\| \\
    & =\sum_{i=1}^{c}\bigg\|\frac{1}{|C_{i}|}(\mathbf{H_{i}^{T}}\mathbf{H_{i}}\mathbf{\Theta}-\mathbf{H_{i}^{T}}\mathbf{Y_{i}})-\frac{1}{|C_{i}^{\prime}|}(\mathbf{H_{i}^{\prime T}}\mathbf{H_{i}^{\prime}}\mathbf{\Theta}-\mathbf{H_{i}^{T \prime}}\mathbf{Y_{i}^{\prime}})\bigg\|^{2} \\
    & =\sum_{i=1}^{c}\bigg\|\frac{1}{|C_{i}|}\mathbf{H_{i}^{T}}\mathbf{H_{i}}\mathbf{\Theta}-\frac{1}{|C_{i}^{\prime}|}\mathbf{H_{i}^{\prime T}}\mathbf{H_{i}^{\prime}}\mathbf{\Theta}-\frac{1}{|C_{i}|}\mathbf{H_{i}^{T}}\mathbf{Y_{i}}+\frac{1}{|C_{i}^{\prime}|}\mathbf{H_{i}^{T \prime}}\mathbf{Y_{i}^{\prime}}\bigg\|^{2} \\
    & \leq \sum_{i=1}^{c}\bigg\|\frac{1}{|C_{i}|}\mathbf{H_{i}^{T}}\mathbf{Y_{i}}-\frac{1}{|C_{i}^{\prime}|}\mathbf{H_{i}^{T \prime}}\mathbf{Y_{i}^{\prime}}\bigg\|^{2}+\sum_{i=1}^{c}\bigg\|\frac{1}{|C_{i}|}\mathbf{H_{i}^{T}}\mathbf{H_{i}}-\frac{1}{|C_{i}^{\prime}|}\mathbf{H_{i}^{\prime T}}\mathbf{H_{i}^{\prime}}\bigg\|^{2}\|\mathbf{\Theta}\|^{2} \\
    & =\sum_{i=1}^{c}\bigg\|\frac{1}{|C_{i}|}\mathbf{Y_{i}^{T}}\mathbf{H_{i}}-\frac{1}{|C_{i}^{\prime}|}\mathbf{Y_{i}^{\prime T}}\mathbf{H_{i}^{\prime}}\bigg\|^{2}+\sum_{i=1}^{c}\bigg\|\frac{1}{|C_{i}|}\mathbf{H_{i}^{T}}\mathbf{H_{i}}-\frac{1}{|C_{i}^{\prime}|}\mathbf{H_{i}^{\prime T}}\mathbf{H_{i}^{\prime}}\bigg\|^{2}\|\mathbf{\Theta}\|^{2}.
    \end{split}
\end{equation}
Here, the first term corresponds to the mean feature alignment and the second term represents feature covariance alignment.
\end{proof}

\subsection{Complexity Analysis}
\label{complexity}

In this part, we give a detailed complexity analysis of three stages: pre-condensation, data transfer, and model retraining. For simplicity, We suppose the model to be unlearned is a GCN and we use $d$ to denote both feature dimension and hidden units of GNNs. $L$ is the number of GNN layers and $\theta_{t}$ denotes the GNN parameters. $r$ is the number of sampled neighbors per node. We use $E$ to present the number of edges for $\mathcal{G}$. $N_{r}, E_{r}$ denote the node number and edge number of the rest of the graph $\mathcal{G}_{r}$.

\vspace{1mm} 
\noindent \textbf{Complexity of pre-condensation.}The complexity of preprocessing for original graph $\mathcal{G}$ is $\mathcal{O}(KEd+KNd^{2})$. During condensation, the inference for condensed graph structure $\mathbf{A^{\prime}}$ has a complexity of $\mathcal{O}(N^{\prime 2}d^{2})$. The complexity of $K$-hop feature propagation for $\mathcal{G^{\prime}}$ is $\mathcal{O}(KN^{\prime 2}d+KN^{\prime}d)$. Compute $\mathcal{L}_{mean}$ has a complexity of $\mathcal{O}(K(N^{\prime}+C)d)$. The complexity of calculating $\mathcal{L}_{cov}$ is $\mathcal{O}(K(N^{\prime}+C)d^{2})$. The complexity of logits alignment is $\mathcal{O}(LN^{\prime 2}d+N^{\prime}(Ld+C))$. Calculation of second-order derivatives in backward propagation has a complexity of $\mathcal{O}(|\theta_{t}|+|\mathbf{A^{\prime}}|+|\mathbf{X^{\prime}}|)$. Considering the epoch number of condensation is $T_{cond}$, the overall complexity can be simplified as $\mathcal{O}(KEd+KNd^{2})+T_{cond} \cdot \mathcal{O}(N^{\prime 2}(d^{2}+Kd+Ld)+KN^{\prime}d^{2})$.

\vspace{1mm} 
\noindent \textbf{Complexity of data transfer.} The complexity of preprocessing for $\mathcal{G}_{r}$ is $\mathcal{O}(KE_{r}d+KN_{r}d^{2}+CN_{r}d)$. During data transfer, the inference for $\mathbf{A^{\prime}}$ has a complexity of $\mathcal{O}(N^{\prime 2}d^{2})$. The complexity of trajectory-based function sampling is $T_{s} \cdot \mathcal{O}(LN^{\prime 2}d+LN^{\prime}d)$. The forward process of GNN is $\mathcal{O}(LN^{\prime 2}d+LN^{\prime}d+r^{L}N_{r}d)$ The complexity of calculating $\mathcal{L}_{feat}$ is $\mathcal{O}(K(N^{\prime}+C)d^{2})$. The complexity of calculating $\mathcal{L}_{sdm}$ is $\mathcal{O}(CN^{\prime}d+C^{2})$. The contrastive-based regularizer $\mathcal{L}_{cdr}$ has a complexity of $\mathcal{O}(N^{\prime 2}d)$. The backward propagation complexity is $\mathcal{O}(|\theta_{t}|+|A^{\prime}|+|X^{\prime}|)$. As the transfer epoch is $T_{ft}$, the overall complexity of the data transfer stage can be simplified as $\mathcal{O}(KE_{r}d+KN_{r}d^{2}+CN_{r}d)+\lfloor \frac{T_{ft}}{\tau_{s}} \rfloor \cdot T_{s}\cdot\mathcal{O}(LN^{\prime 2}d+LN^{\prime}d+r^{L}N_{r}d)+T_{ft} \cdot \mathcal{O}(N^{\prime 2}d^{2}+LN^{\prime 2}d+KN^{\prime}d^{2}+CN^{\prime}d)$.

\vspace{1mm} 
\noindent \textbf{Complexity of model retraining.} The overall complexity of GNN training on the condensed graph is $T_{rt}\mathcal{O}(LN^{\prime 2}d+LN^{\prime}d)$, where $T_{rt}$ is the number of retraining epoch.

\section{Supplementary Experiments}
\label{supply}

\subsection{Evaluation on more GNN Backbones}

To validate the generalization ability of our TCGU, we further employ two widely-used GNN backbones SGC and GIN in node unlearning evaluation. As shown in Table~\ref{backbone}, TCGU consistently achieves SOTA performance in terms of F1-score on three datasets, compared with advanced GU strategies. In particular, we observe that with SGC, TCGU outperforms Retrain by 2.12\% and 1.68\% on Cora and Computers, respectively. For GIN, compared with Retrain, TCGU achieves a performance gain of about 1.48\% and 2.08\% on Cora and Computers, respectively. These results illustrate the superior model utility when unlearning with TCGU. For unlearning efficiency, we find TCGU also achieves comparable running time to approximate methods and exhibits significant efficiency improvement compared to exact unlearning methods with SGC and GIN. Additionally, we notice that unlearning time on SGC is lower than those on the other three backbones, this is mainly because its linear architecture can speed up GNN training, reducing the time cost of our function sampling process and retraining stage. Overall, our TCGU can be applied to unlearn various GNN models and achieve a better trade-off between model utility and unlearning efficiency.

\begin{table}
\caption{Additional GNN backbones}
\label{backbone}
\centering
\resizebox{\linewidth}{!}{\begin{tabular}{cccccccc}
\toprule
\multirow{2}{*}{GNN} & \multicolumn{1}{c|}{\multirow{2}{*}{Component}} & \multicolumn{2}{c}{Cora} & \multicolumn{2}{c}{Citeseer} & \multicolumn{2}{c}{Computers} \\
                     & \multicolumn{1}{c|}{}                        & F1        & RT  & F1        & RT & F1        & RT      \\ \midrule
\multirow{9}{*}{SGC}    &    \multicolumn{0}{c|}{Retrain}              & 81.33\(\pm\)0.47 & 10.74 & 75.54\(\pm\)0.52 & 23.46 & 80.14\(\pm\)0.32 & 140.75  \\
                        &    \multicolumn{0}{c|}{GE-BLPA}              & 67.47\(\pm\)0.02 &13.60 & 68.72\(\pm\)0.01 &14.62 & 72.69\(\pm\)0.01 & 63.40   \\
                        &    \multicolumn{0}{c|}{GE-BKEM}              & 44.18\(\pm\)0.03 & 12.07 & 53.23\(\pm\)0.00 &14.09 & 70.56\(\pm\)0.00 & 62.59    \\
                        &    \multicolumn{0}{c|}{GUIDE-Fast}           & 60.50\(\pm\)0.01 &8.95 & 57.25\(\pm\)0.01 &9.32 & 67.27\(\pm\)0.00 & 52.88    \\
                        &    \multicolumn{0}{c|}{GUIDE-SR}             & 63.99\(\pm\)0.04 & 9.30& 58.10\(\pm\)0.02 &9.48 & 71.53\(\pm\)0.02 &  53.56   \\
                        &    \multicolumn{0}{c|}{GNNDelete}            & 77.67\(\pm\)0.25 & 0.95& 72.82\(\pm\)0.13 & 1.15& 80.02\(\pm\)0.15 &  3.24    \\
                        &    \multicolumn{0}{c|}{GIF}                  & \underline{81.07\(\pm\)0.28} & \textbf{0.38} & \underline{74.79\(\pm\)0.26} & \textbf{0.27} &\underline{81.16\(\pm\)0.33} & \textbf{0.46}     \\
                              &    \multicolumn{0}{c|}{MEGU}                 & 80.59\(\pm\)0.12& \underline{0.68} & 74.68\(\pm\)0.05& 1.01& 80.65\(\pm\)0.17 &4.27     \\
                              &    \multicolumn{0}{c|}{TCGU}                        & \textbf{83.45\(\pm\)0.19} & 0.61 & \textbf{75.29\(\pm\)0.43} & \underline{0.96} & \textbf{81.82\(\pm\)0.36} &  \underline{2.86}   \\ \midrule
\multirow{9}{*}{GIN}          &    \multicolumn{0}{c|}{Retrain}                         &80.41\(\pm\)0.44 & 11.65 &74.23\(\pm\)0.27 &20.96 &81.34\(\pm\)0.45 & 123.28  \\
                              &    \multicolumn{0}{c|}{GE-BLPA}                        &64.38\(\pm\)0.05 & 10.13& 62.44\(\pm\)0.05& 11.83&75.51\(\pm\)0.04 & 54.74    \\
                              &    \multicolumn{0}{c|}{GE-BKEM}                        &38.45\(\pm\)0.03 & 9.82& 55.89\(\pm\)0.03&11.46 & 66.07\(\pm\)0.02& 54.78    \\
                              &    \multicolumn{0}{c|}{GUIDE-Fast}                         & 67.26\(\pm\)0.06&6.74 & 69.25\(\pm\)0.02&7.35 & 73.06\(\pm\)0.05& 43.92    \\
                              &    \multicolumn{0}{c|}{GUIDE-SR}                        & 68.39\(\pm\)0.02& 6.88& 69.40\(\pm\)0.05& 8.76& 73.93\(\pm\)0.03 & 45.11    \\
                              &    \multicolumn{0}{c|}{GNNDelete}                        &74.21\(\pm\)0.25 & 1.29 & 72.44\(\pm\)0.29 & 1.72 & 77.29\(\pm\)0.21 & 3.62     \\
                              &    \multicolumn{0}{c|}{GIF}                       & 72.70\(\pm\)0.38 & \textbf{0.32} & 73.59\(\pm\)0.26& \textbf{0.35} & 77.53\(\pm\)0.33 & \textbf{0.98}    \\
                              &    \multicolumn{0}{c|}{MEGU}                       & \underline{78.82\(\pm\)0.08} & \underline{0.58} & \textbf{75.19\(\pm\)0.15} & \underline{0.65} & \underline{81.78\(\pm\)0.09}& \underline{2.39}   \\
                              &    \multicolumn{0}{c|}{TCGU}                        & \textbf{81.89\(\pm\)0.26}  & 1.94 & \underline{74.35\(\pm\)0.41} & 2.37 & \textbf{83.42\(\pm\)0.45} & 4.22     \\ \bottomrule
\end{tabular}}
\end{table}

\subsection{Evaluation on more Unlearning Tasks}

\begin{figure}[t]
	\centering
	\subfloat[Cora (GCN)]{
		\includegraphics[width=0.45\linewidth]{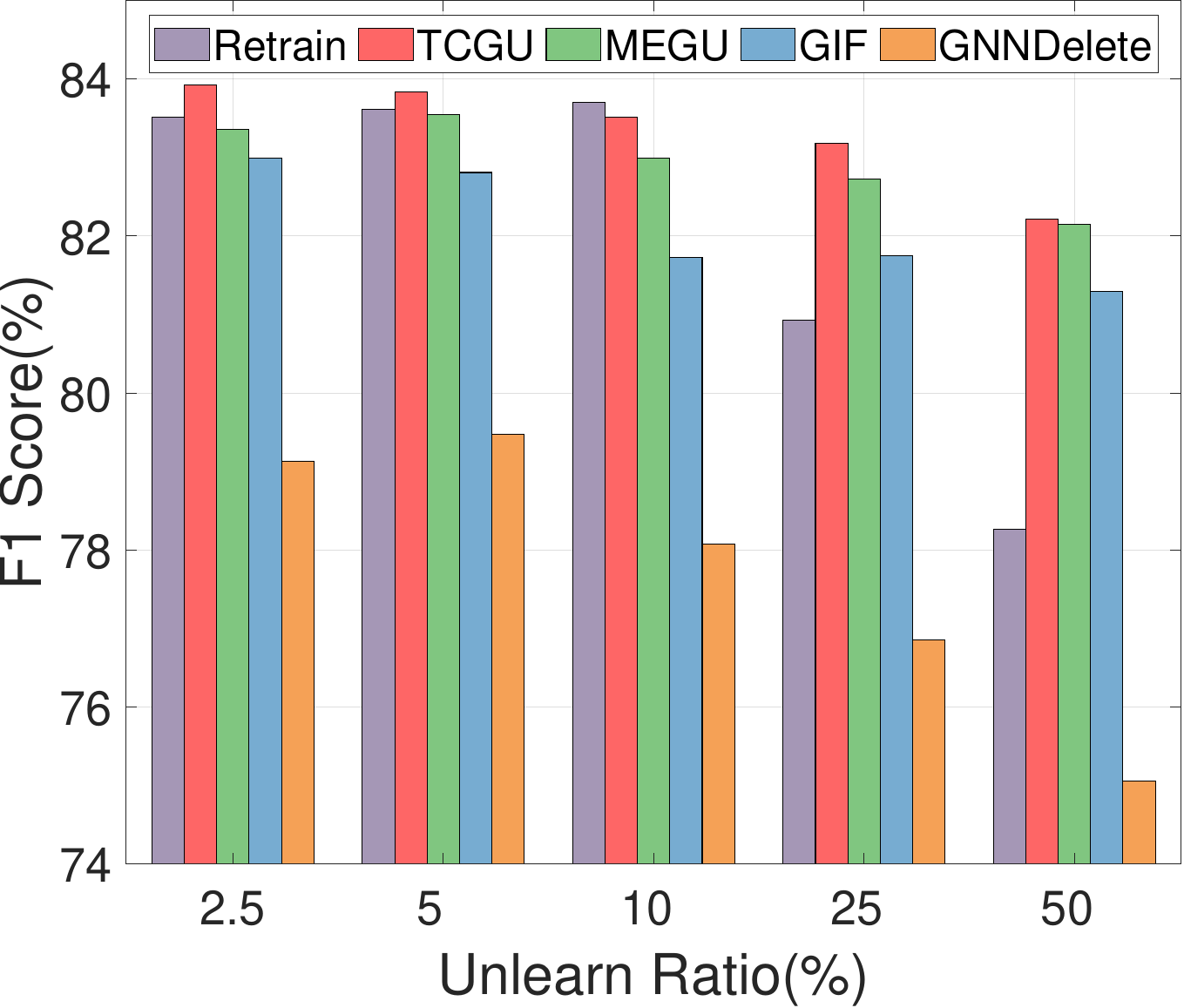}}
        \subfloat[Cora (GAT)]{
		\includegraphics[width=0.45\linewidth]{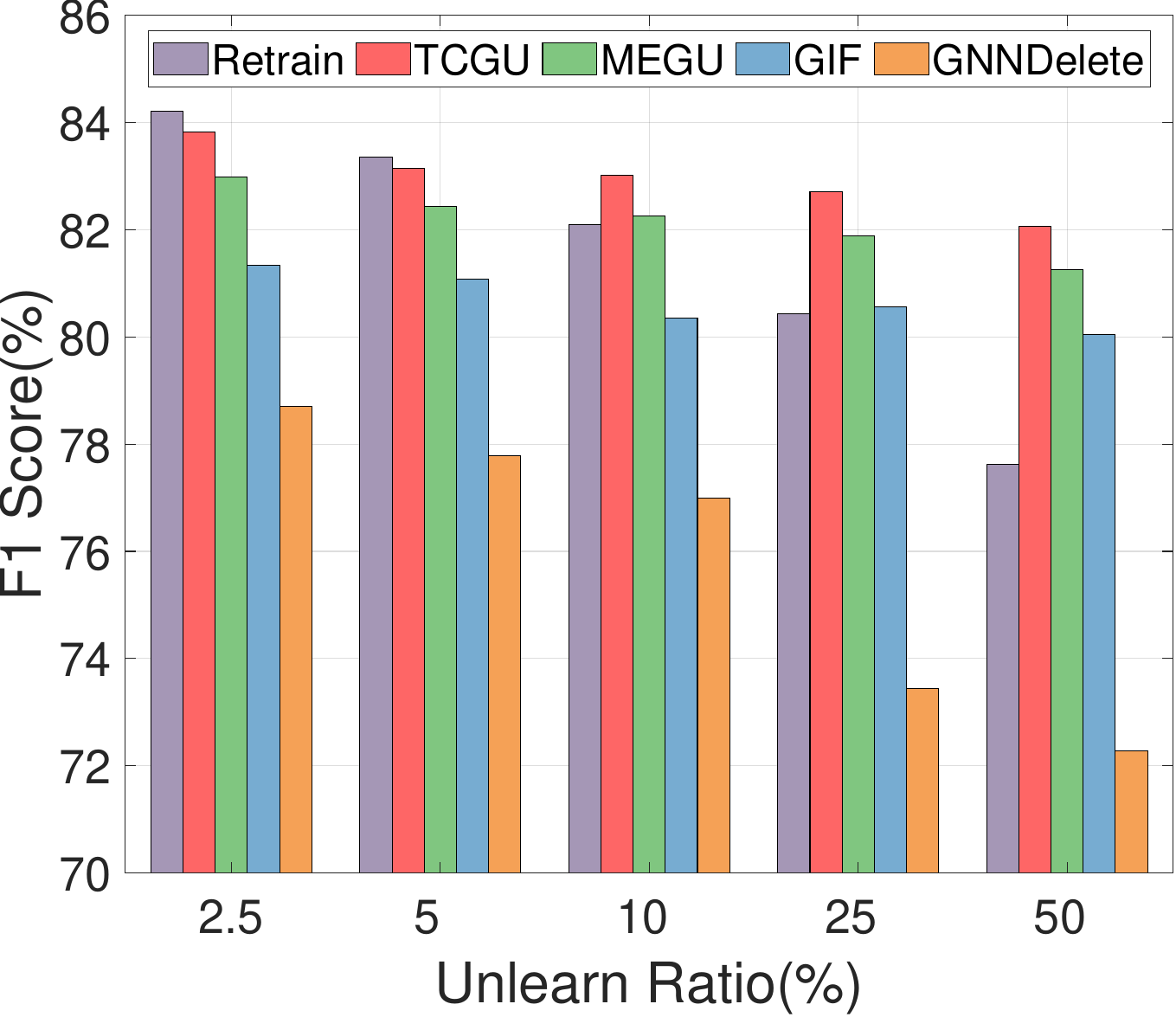} }
	\\
	\subfloat[Photo (GCN)]{
		\includegraphics[width=0.45\linewidth]{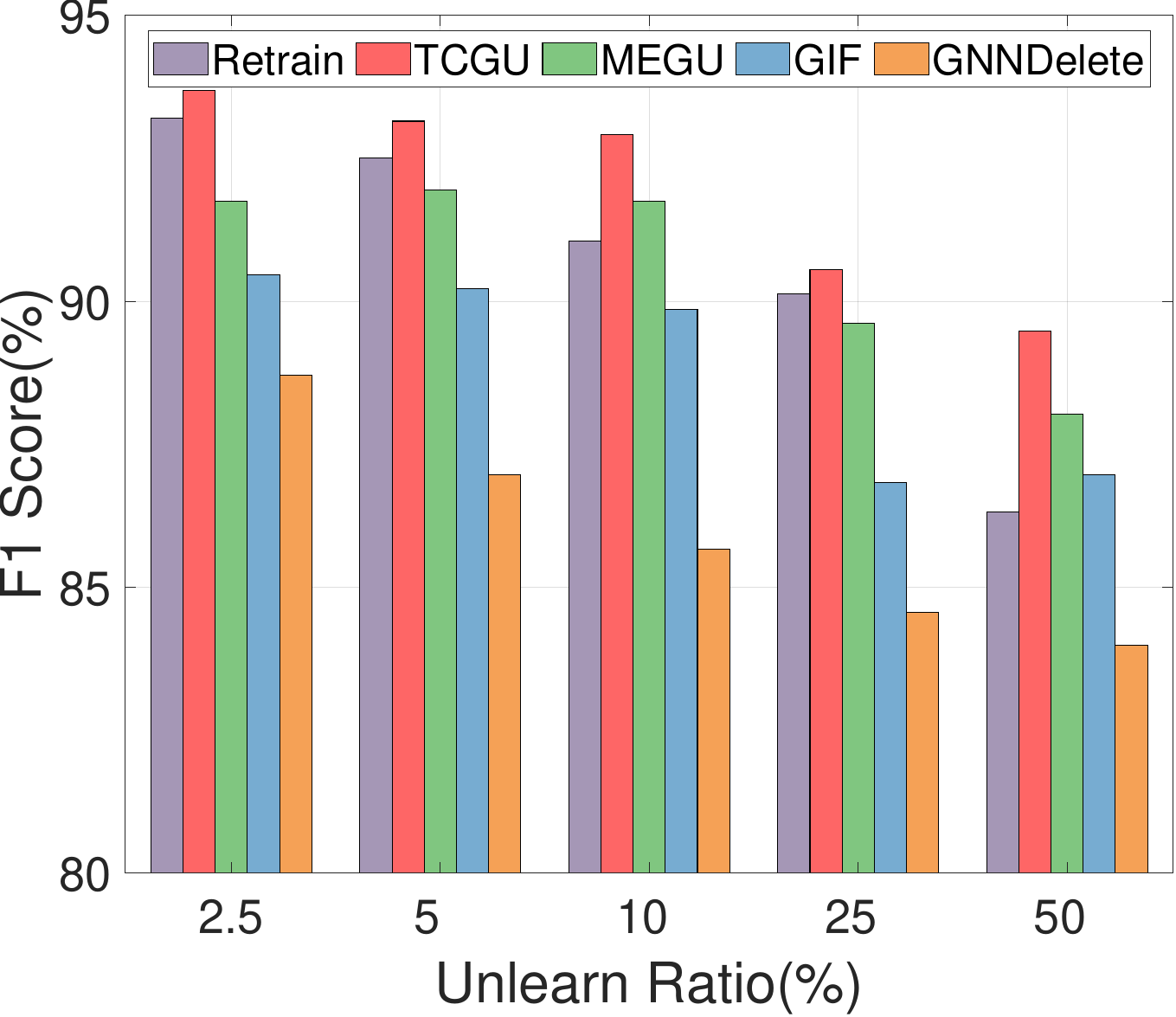}}
	\subfloat[Photo (GAT)]{
		\includegraphics[width=0.45\linewidth]{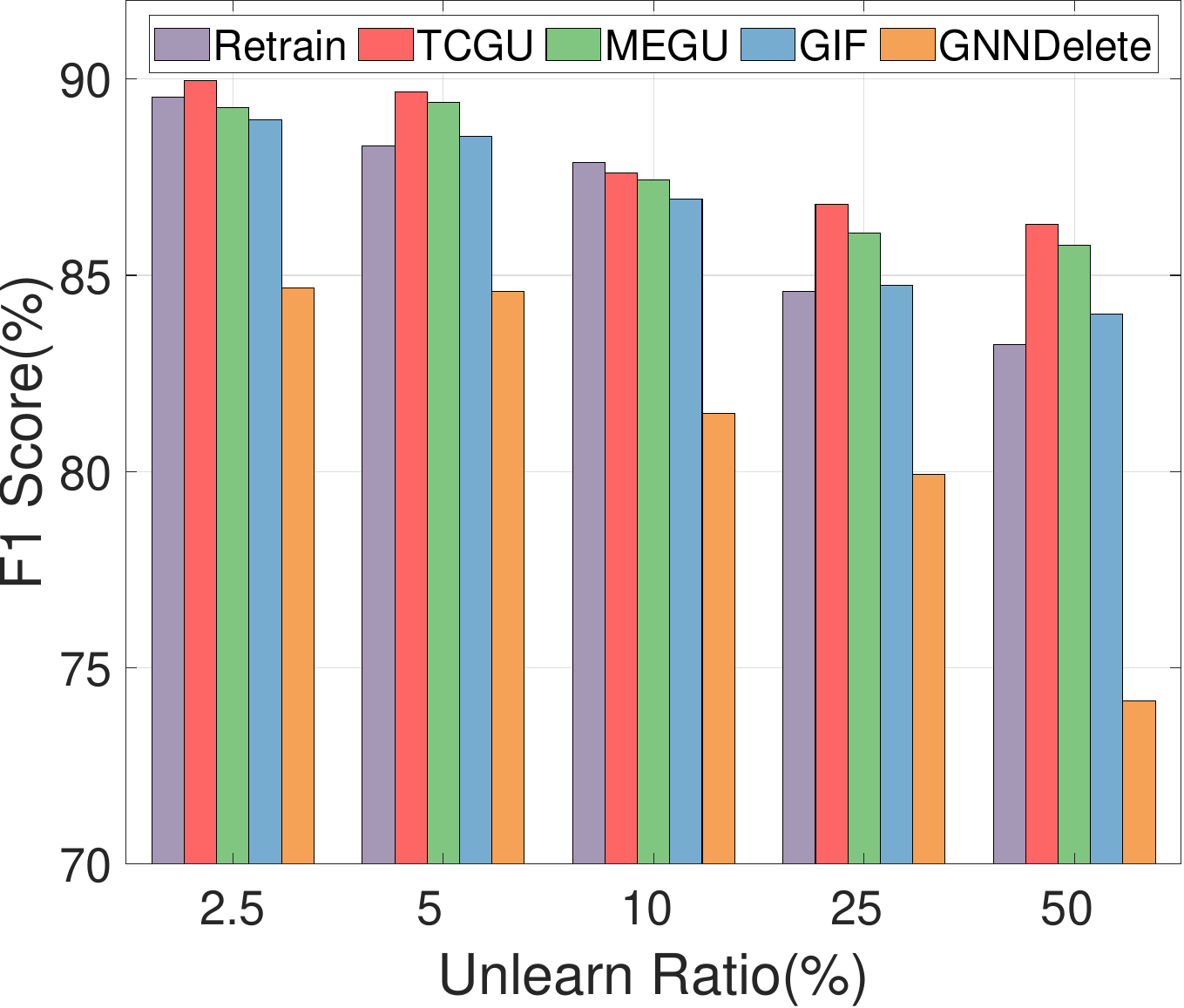}}
        \caption{Edge unlearning with different ratios}
        \label{fig:edgeunlearn}
\end{figure}

\begin{figure}[t]
	\centering
	\subfloat[Cora (GCN)]{
		\includegraphics[width=0.45\linewidth]{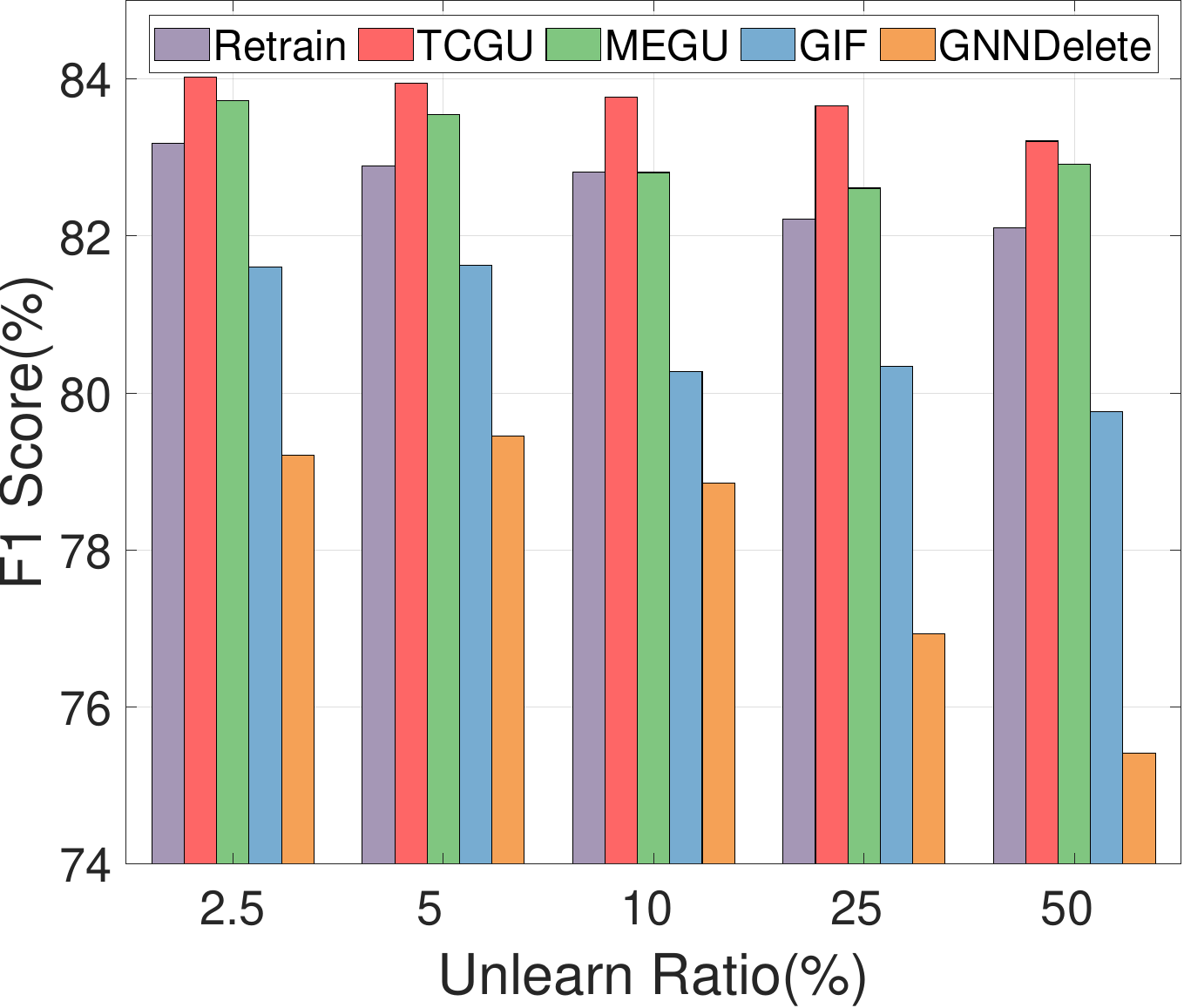}}
	\subfloat[Cora (GAT)]{
		\includegraphics[width=0.45\linewidth]{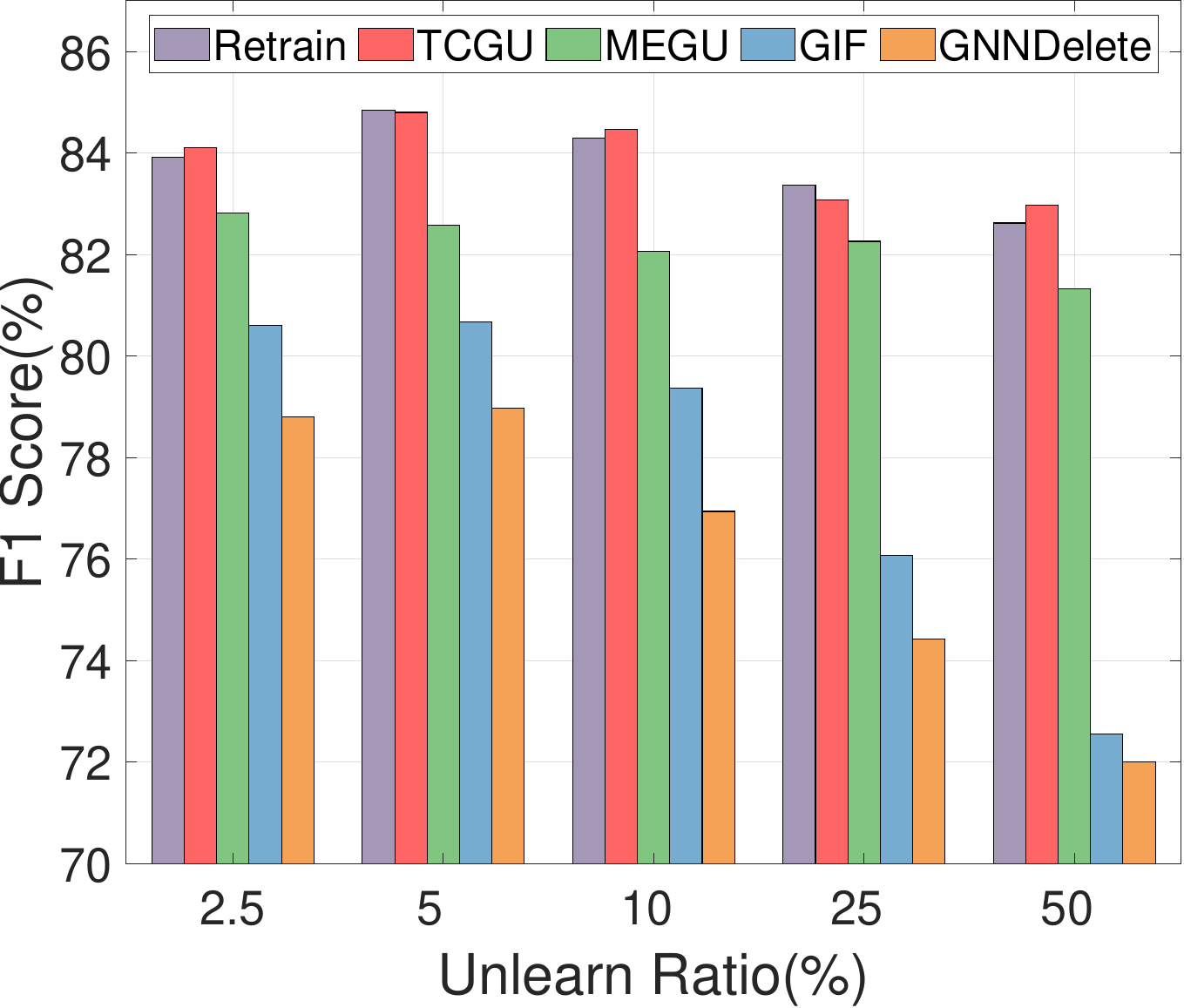}}
        \\
        \subfloat[Photo (GCN)]{
		\includegraphics[width=0.45\linewidth]{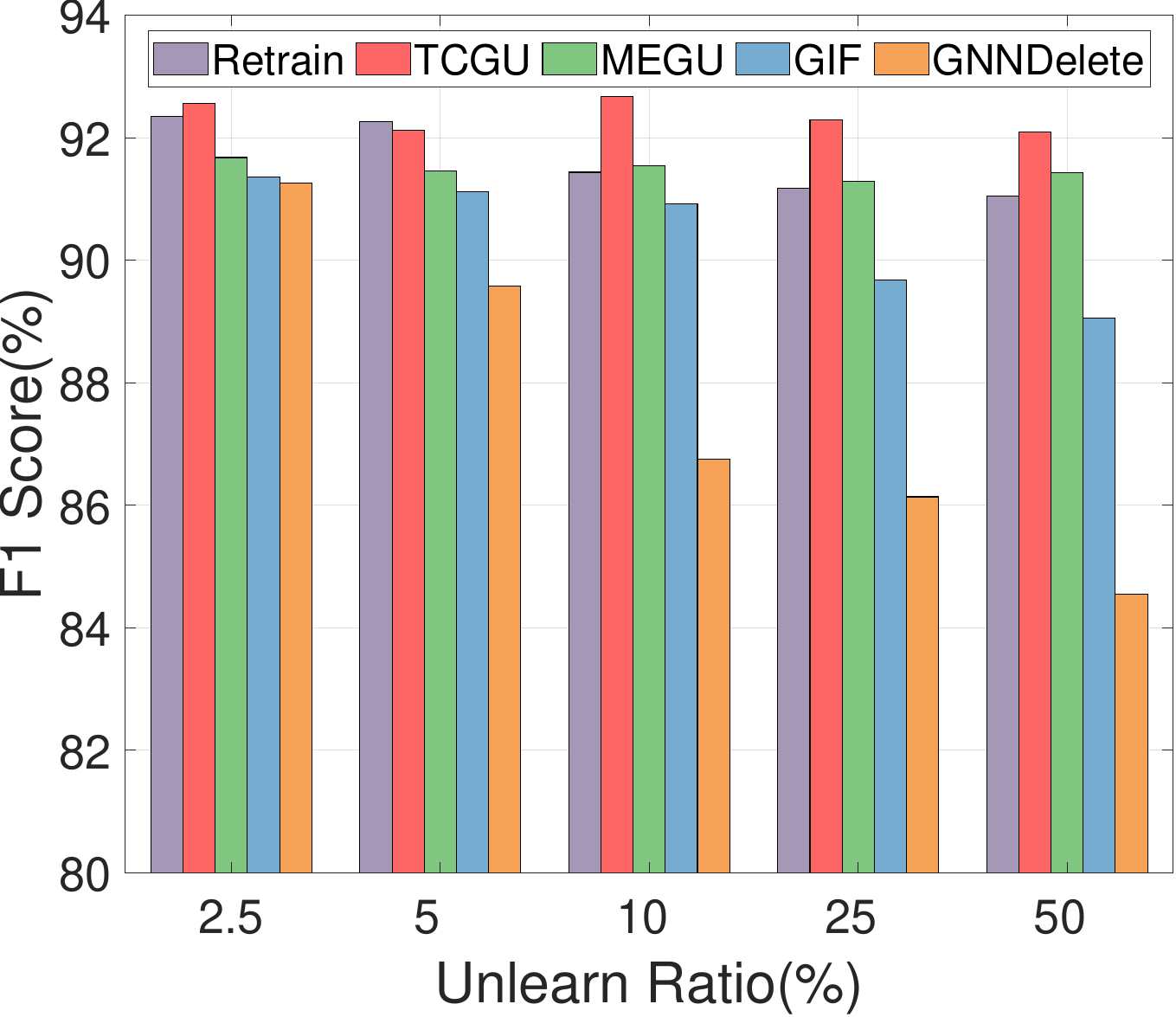}}
        \subfloat[Photo (GAT)]{
		\includegraphics[width=0.45\linewidth]{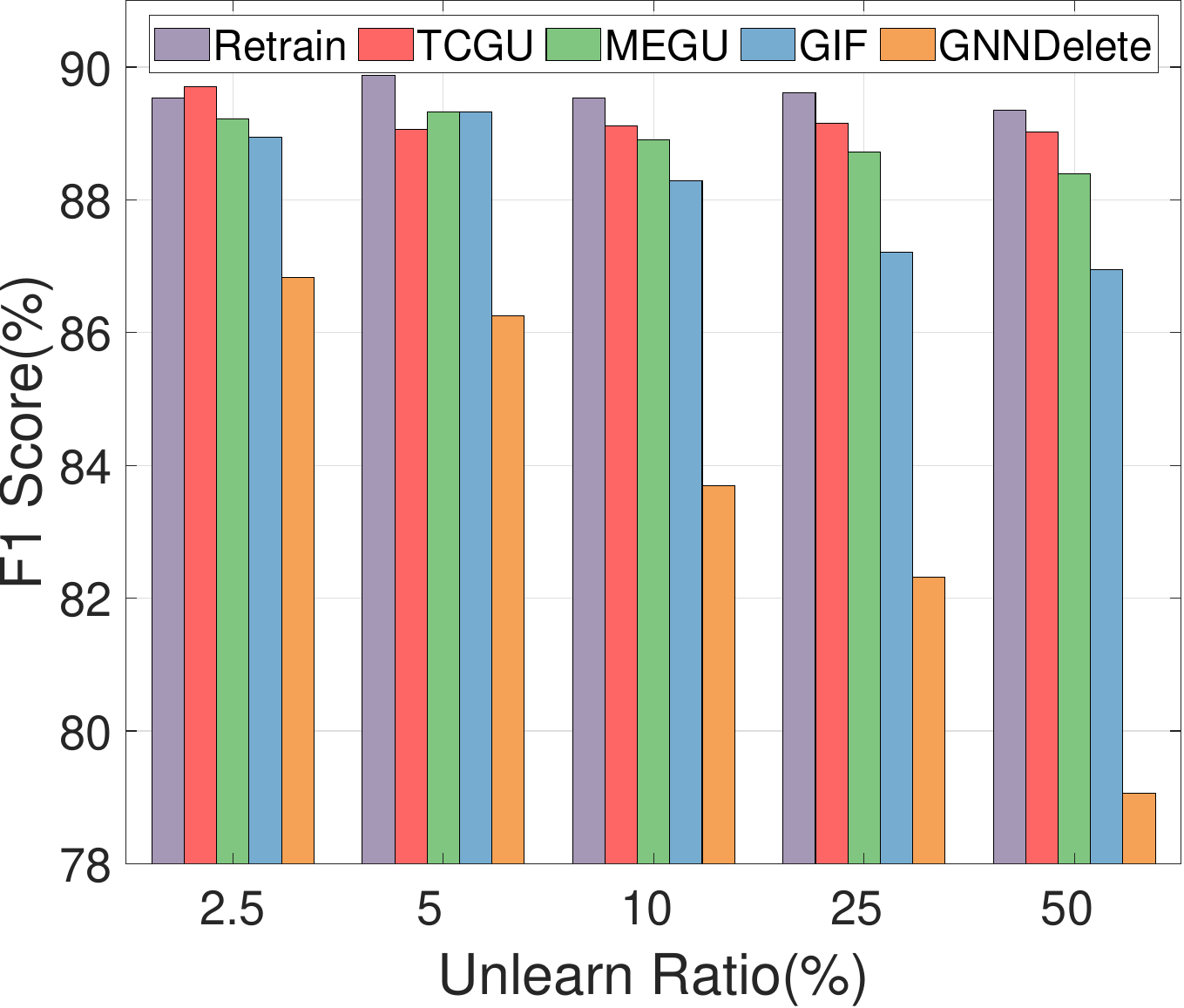}}
        \caption{Feature unlearning with different ratios}
        \label{fig:featunlearn}
\end{figure}

To evaluate TCGU on different unlearning tasks, we 
conduct edge unlearning and feature unlearning with different unlearning ratios \(\rho\%\) on the Cora and Photo datasets with two representative GNN models, GCN and GAT. The experimental results are illustrated in Figure~\ref{fig:edgeunlearn} and Figure~\ref{fig:featunlearn}, respectively.

In general, TCGU consistently achieves comparable or even superior performance to the Retrain method on both edge and feature unlearning tasks. Besides, it also outperforms existing approximate strategies in different cases. For example, on the Cora dataset with the GCN model, when the ratio of unlearned edges increases from 10\% to 25\%, the performance of Retrain drops significantly from 0.837 to 0.809. In contrast, TCGU only suffers a small performance drop from 0.835 to 0.831. Using GCN on the Photo dataset, TCGU is the only model that achieves over 0.92 F1-score under different feature unlearning ratios. These results validate our TCGU can preserve model utility under different unlearning requests. We also observe that for feature unlearning tasks, model performance is less sensitive to the unlearning scale compared to edge unlearning. We analyze this may be because the deletion of edges impairs graph topology which is more significant than node attributes in graph representation learning, leading to a larger performance drop. The feature unlearning which masks features of some nodes would cause a relatively milder impact on model performance.

\subsection{Parameter Sensitivity Analysis}

\begin{figure}[t]
	\centering
	\subfloat[Cora]{
		\includegraphics[width=0.32\linewidth]{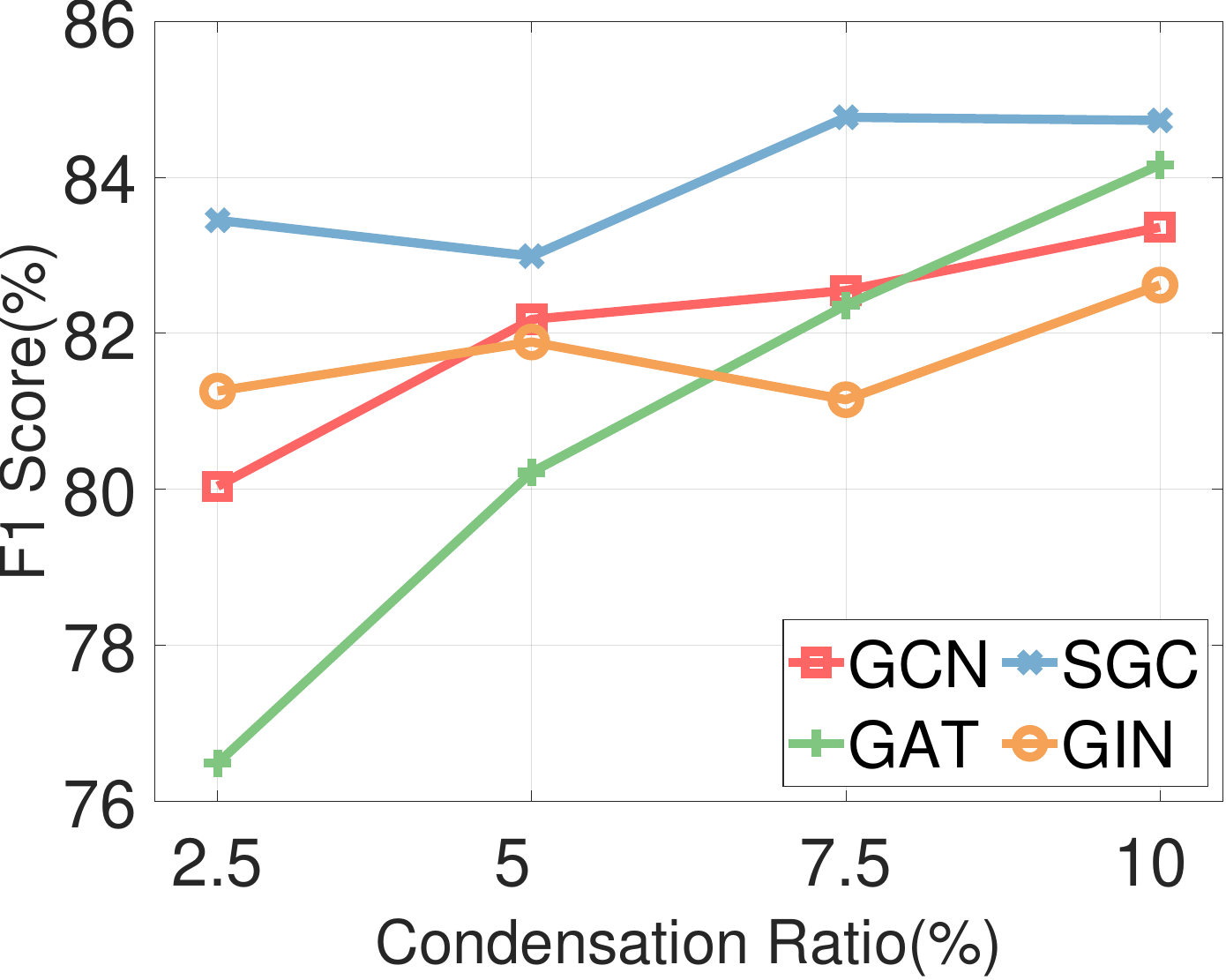}}
	\subfloat[Photo]{
		\includegraphics[width=0.32\linewidth]{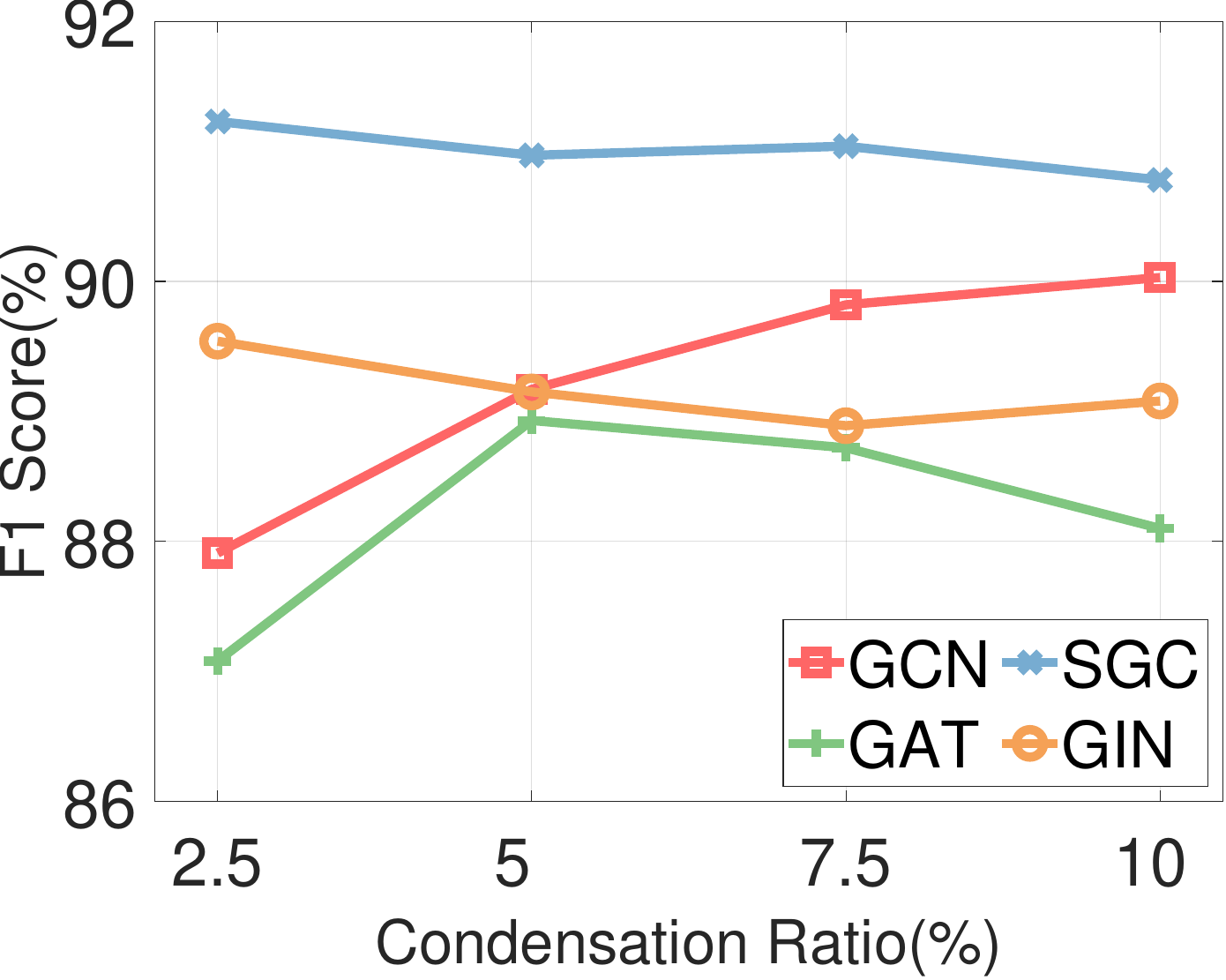}}
        \subfloat[Computers]{
		\includegraphics[width=0.32\linewidth]{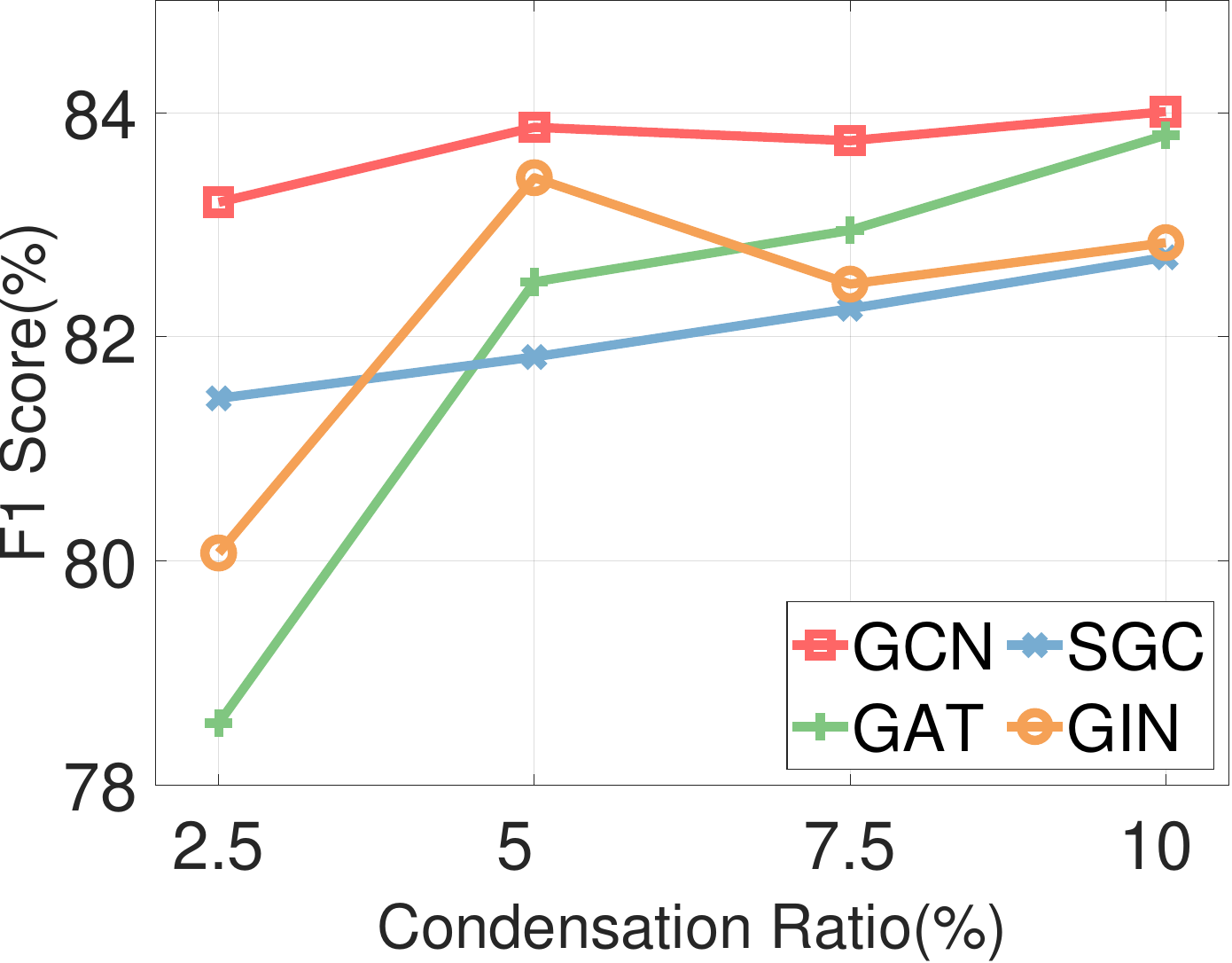}}
        \caption{Parameter sensitivity on condensation ratio \(r_{cond}\)}
        \label{fig:cond_ratio}
\end{figure}

\textbf{Condensation Ratio \(r_{cond}\).} We further study the effect of condensation ratio \(r_{cond}\) on model utility. In detail, we vary \(r_{cond}\) in a range of \(\{2.5\%,5\%,7.5\%,10\%\}\) for node unlearning on Cora, Photo, and Computers with 4 GNNs. The results are reported in Figure~\ref{fig:cond_ratio}. From the Figure, we can observe that the increasing \(r_{cond}\) can boost model performance in general. This is because larger condensed data can capture more patterns in the original graph. Notably, we find that GAT is more sensitive to the change of \(r_{cond}\) than the other 3 backbones. This is because GAT relies on
edge-based attention mechanisms for information aggregation. As \(r_{cond}\) increases, a more evident change in graph topology will occur, significantly affecting the performance of GAT.

\begin{table}[th]
    \centering
    \caption{Analysis on \(\lambda_{f}^{\prime},\lambda_{r}^{\prime}\). We report averaged F1-score} 
    \label{param_tune}
    \begin{subtable}{0.8\linewidth}
      \centering
        \caption{Cora}
        \resizebox{\linewidth}{!}{
        \begin{tabular}{cccccc}
\toprule
\(\lambda_{f}^{\prime} \backslash \lambda_{r}^{\prime}\) & \(5e^{-4}\) & \(1e^{-3}\) & \(2e^{-3}\) & \(5e^{-3}\)  \\ \midrule
50 & 76.89&	81.26&	81.89&	80.33     \\
100& 76.93&	81.99 & \textbf{82.18} & 80.52     \\ 
150  & 75.97 & 80.96 & 81.15 & 78.74        \\ 
200 & 75.05	& 82.07	& 79.68	& 78.93   \\ \bottomrule
\end{tabular}
        }
    \end{subtable}%
    \\
    \begin{subtable}{0.8\linewidth}
      \centering
        \caption{Computers}
        \resizebox{\linewidth}{!}{
\begin{tabular}{cccccc}
\toprule
\(\lambda_{f}^{\prime} \backslash \lambda_{r}^{\prime}\) & \(5e^{-4}\) & \(1e^{-3}\) & \(2e^{-3}\) & \(5e^{-3}\)  \\ \midrule
50  & 79.48 & 82.42 & 81.71 & 80.25       \\
100 & 82.69 & 83.75 & 82.62 & 80.33     \\ 
150 & 83.80	& \textbf{84.76} & 83.27 & 81.31        \\ 
200 & 83.22	& 84.55	& 81.75 & 80.84    \\ \bottomrule
\end{tabular}
        }
    \end{subtable} 
\end{table}

\vspace{1mm} 
\noindent \textbf{Transfer Coefficients \(\lambda_{f}^{\prime},\lambda_{r}^{\prime}\).} In this subsection, we investigate the effect of \(\lambda_{f}^{\prime},\lambda_{r}^{\prime}\) in the fine-tuning stage. We vary \(\lambda_{f}^{\prime}\) in a range of \(\{50,100,150,200\}\) and \(\lambda_{r}^{\prime}\) in a range of \(\{5e^{-4},1e^{-3},2e^{-3},5e^{-3}\}\) and conduct node unlearning task on Cora and Computers with GCN backbone. Averaged results are reported in Table~\ref{param_tune}. As can be seen, the optimal combination of \(\lambda_{f}^{\prime},\lambda_{r}^{\prime}\) is different in two datasets. For example, on the Cora dataset, a larger \(\lambda_{r}^{\prime}\) (\(2e^{-3}\)) leads to better utility, while for Computers, a small regularization weight (\(5e^{-4}\)) can achieve satisfactory performance. We also find that when one coefficient is fixed and increases the other, the model utility usually increases first and then gradually degrades. This shows that over-imbalance on different loss terms may result in a performance drop.

\end{document}